\documentclass{article}

\PassOptionsToPackage{numbers, compress}{natbib}

\usepackage[preprint]{neurips_2024}

\usepackage[utf8]{inputenc} 
\usepackage[T1]{fontenc}    
\usepackage{hyperref}       
\usepackage{url}            
\usepackage{booktabs}       
\usepackage[table]{xcolor}
\usepackage{amsfonts}       
\usepackage{nicefrac}       
\usepackage{microtype}      
\usepackage{xcolor}         
\usepackage{multirow}
\usepackage{array}
\usepackage{graphicx} 
\usepackage{amsmath} 
\usepackage{tcolorbox}
\usepackage{booktabs}
\usepackage{subcaption}
\usepackage{arydshln}
\usepackage{appendix} 
\usepackage{titletoc}
\usepackage{floatflt}
\usepackage{paralist}
\usepackage[normalem]{ulem}

\colorlet{pale1}{blue!10}
\colorlet{pale2}{green!10}
\colorlet{pale3}{red!10}
\colorlet{pale4}{orange!10}
\colorlet{pale5}{cyan!10}
\colorlet{pale6}{magenta!10}
\colorlet{pale7}{gray!10}
\colorlet{pale8}{teal!10}
\colorlet{pale9}{purple!10}
         
\cellcolor{pale7}

\title{AdaptAgent: Adapting Multimodal Web Agents with Few-Shot Learning from Human Demonstrations}

\author{%
Gaurav Verma$^1$ \quad Rachneet Kaur$^2$ \quad Nishan Srishankar$^2$\\ \textbf{Zhen Zeng}$^2$ \quad
\textbf{Tucker Balch}$^2$ \quad \textbf{Manuela Veloso}$^2$\\
$^1$Georgia Institute of Technology \quad\quad $^2$J.P. Morgan AI Research\\
\texttt{gverma@gatech.edu} \quad \texttt{\{rachneet.kaur, nishan.srishankar\}@jpmorgan.com} \\
\texttt{\{zhen.zeng, tucker.balch, manuela.veloso\}@jpmorgan.com}\\
}

\begin{document}

\maketitle

\begin{abstract}

State-of-the-art multimodal web agents, powered by Multimodal Large Language Models (MLLMs), can {autonomously} execute many web tasks by processing user instructions and {interacting with graphical user interfaces (GUIs)}. Current strategies for building web agents rely on \textit{(i)} the generalizability of underlying MLLMs and their steerability via prompting, and \textit{(ii)} large-scale fine-tuning of MLLMs on web-related tasks. However, web agents still struggle to automate tasks on unseen websites and domains, limiting their applicability to enterprise-specific and proprietary platforms. Beyond generalization from large-scale pre-training and fine-tuning, we propose building agents for few-shot adaptability {using} human demonstrations. We introduce the \texttt{AdaptAgent} framework that enables both proprietary and open-weights multimodal web agents to adapt to new websites and domains using few human demonstrations (up to 2). Our experiments on two popular benchmarks — Mind2Web \& VisualWebArena — show that using in-context demonstrations (for proprietary models) or meta-adaptation demonstrations (for meta-learned open-weights models) boosts task success rate by 3.36\% to 7.21\% over non-adapted state-of-the-art models, corresponding to a relative increase of 21.03\% to 65.75\%. Furthermore, our additional analyses \textit{(a)} show the effectiveness of multimodal demonstrations over text-only ones, \textit{(b)} shed light on the influence of different data selection strategies during meta-learning on the generalization of the agent, and \textit{(c)} demonstrate the effect of number of few-shot examples on the web agent's success rate. Overall, our results unlock a complementary axis for developing widely applicable multimodal web agents beyond large-scale pre-training and fine-tuning, emphasizing few-shot adaptability.
\end{abstract}

\section{Introduction}

Agents automating web-based tasks with minimal human {intervention} can {significantly} boost personal and workplace productivity~\cite{noy2023experimental, oracle_ai_agents_2024}. A prevalent interaction mechanism involves a human providing a natural language instruction (e.g., \textit{``use delta.com to book a flight from JFK to Haneda on …’’}), and the agent {autonomously} executing the necessary webpage actions {to complete the user-assigned task}~\cite{zheng2024seeact, deng2023mind2web, hong2023cogagent}. Large language models (LLMs) can understand instructions, plan, and predict structured outputs, serving as backbones for such agents~\cite{veloso2005perception}. Remarkable progress has been made in automating web-based tasks using LLM-based agents~\cite{lai2024autowebglm, cheng2024seeclick, he2024webvoyager}, employing careful prompting~\cite{zheng2024seeact, koh2024visualwebarena} and extensive pre-training and fine-tuning~\cite{deng2023mind2web} to predict actions using language instructions and HTML/{DOM}. With multimodal capabilities, these agents now process the {graphical user interface's (GUI's)} visual state to complement the HTML/DOM information~\cite{hong2023cogagent}. In parallel with the methodological advancements, evaluating {the generalizability of these multimodal web agents} to new tasks, websites, and domains {is} a critical component to ensure their broad applicability.

Prior works have noted challenges in generalizing multimodal web agents to new tasks, websites, and domains, while often relying on large-scale pre-training (e.g., agents like SeeAct~\cite{zheng2024seeact}) or fine-tuning (e.g., models like CogAgent~\cite{hong2023cogagent}). We posit that regardless of pre-training scale, some tasks and domains will remain unseen{, such as} proprietary workflows and enterprise websites. Since {the} generalizability of current state-of-the-art {(SoTA)} agents is {limited} and their fine-tuning is costly, we propose \textit{building web-agents for data-efficient adaptability} instead of relying solely on large-scale pre-training and fine-tuning. Specifically, we address whether multimodal web agents can adapt to unseen websites and domains with only a handful of human demonstrations ({e.g.,} $n = 1$ or $n = 2$).

We consider current {SoTA} multimodal web agents — both proprietary and open-weights — and demonstrate that incorporating just 1 or 2 multimodal human demonstrations (visual snapshot + HTML information) can {result in an absolute increase in task success rate of 3.36\% to 7.21\%} on unseen websites and domains, corresponding to a relative increase of  21.03\% to 65.75\% over current performance.
We propose the \texttt{AdaptAgent} framework to effectively use these few-shot demonstrations {through} careful in-context learning (ICL)~\cite{brown2020language} with proprietary multimodal LLMs (MLLMs) and meta-learning~\cite{finn2017one} with open-weights multimodal LLMs. To establish the role of learning from few-shot demonstrations, we conduct extensive experiments on two widely adopted benchmarks --- Mind2Web~\cite{deng2023mind2web} \& VisualWebArena~\cite{koh2024visualwebarena} {---} showing improvements across tasks of varying difficulty levels.  Our key contributions are summarized as:

\noindent\textbullet \hspace{1pt} We propose the \texttt{AdaptAgent} framework for enabling {SoTA} multimodal web agents to learn from few-shot human demonstrations. \texttt{AdaptAgent} uses ICL for data-efficient adaptation of proprietary MLLMs like GPT-4o~\cite{achiam2023gpt} and meta-learning for adapting open-weights MLLMs like CogAgent~\cite{hong2023cogagent}.\\
\noindent\textbullet \hspace{1pt} Our extensive experiments on Mind2Web and VisualWebArena demonstrate the effectiveness of our methods, resulting in notable increases in task success rates on unseen websites and domains with only 1 or 2 multimodal demonstrations.\footnote{{For a more granular investigation of the observations, we conduct ablations to break down the main results, stratifying improvements based on action sequence complexity and visual difficulty. See Appendix \ref{app_sec:difficulty-level-results}.}}\\
\noindent\textbullet \hspace{1pt} We conduct additional analyses that provide actionable recommendations for future work by researchers and practitioners. Specifically, we show that \textit{(a)} multimodal in-context demonstrations are more effective than text-only demonstrations, \textit{(b)} different data selection strategies for meta-learning influence the post-adaptation generalization of the adapted agent, and \textit{(c)} more demonstrations help in boosting agent's performance, but correspond to higher computational costs and saturating gains.

We believe that the effectiveness of using few-shot human demonstrations and our empirical insights open a complementary direction for improving the generalizability of multimodal web agents beyond {the current SoTA} strategies that rely on large-scale pre-training and fine-tuning.

\section{Related Work}

We categorize the related prior work along three dimensions: work on UI/Web agents, few-shot learning approaches with LLMs, and approaches to learn from demonstrations. An expanded discussion of the prior work is presented in Appendix \ref{sec:detailed_related_work}.

\textbf{UI/Web Agents}: Controlling digital devices using AI and natural language input has been a long-standing goal~\cite{shi2017world, humphreys2022data}. Before large language models (LLMs), approaches often used reinforcement learning on top of models like LSTM and BERT for language processing, combined with ResNet-like models for GUI state understanding~\cite{liu2018reinforcement, iki2022berts}. With the advent of multimodal LLMs, recent work has leveraged these models to build web agents that process user instructions and reason to generate actions on user interfaces~\cite{zheng2024seeact, he2024webvoyager}.

Most state-of-the-art methods use pretrained LLMs, such as GPT-4, to build multimodal web agents. They provide the LLM with context like images of the GUI, prior actions, image annotations, and HTML/DOM information when available. Some works, like Pix2Act~\cite{shaw2023pixels} and WebAgent~\cite{gur24webagent}, train LLMs to attend to parts of HTML code or generate the next action step through self-supervision, often using reinforcement learning techniques like behavioral cloning or REINFORCE. However, these approaches typically require large amounts of training data and resources, and are often limited to simpler environments~\cite{lai2024autowebglm}. They may not scale well to complex proprietary enterprise software, and agents requiring exploration during training may need human supervision to avoid risky outcomes. Methods that aim to make agents more adaptable to unseen settings, which is the focus of this work, could avoid costly retraining processes, enhance applicability to proprietary settings, and allow agents to learn from custom information provided by human experts. Related to the theme of unlocking new agent capabilities, recent work has investigated giving web agents access to APIs~\cite{song2024beyond}, mapping large-scale indirect knowledge to supervision  signals for improving agent's performance~\cite{ou2024synatra}, and carefully constructing reasoning-based benchmark of tasks for facilitating comprehensive evaluations~\cite{boisvert2024workarena}. 

\textbf{Few-Shot Learning with LLMs}: Data-efficient alignment of LLMs to new tasks is an active area of research~\cite{jin2023dataefficientalignmentlargelanguage, liu2024deita}. While in-context learning~\cite{brown2020language} allows models to adapt using few examples, it can be sensitive to variations~\cite{khattab2023dspy, sclar2023quantifying}. Fine-tuning methods like Group Preference Optimization (GPO)~\cite{zhao2023group} and DITTO~\cite{shaikh2024show} have shown promise in few-shot tuning of LLMs to align with subjective preferences. However, these methods are designed for preference tuning and may not directly translate to tasks requiring precise action prediction. Inspired by the potential of meta-learning, we adopt model-agnostic meta-learning~\cite{finn2017modelfirst} to train web agents that can quickly adapt using few-shot demonstrations. This approach aims to improve the performance of multimodal web agents, especially in cross-website and cross-domain scenarios.

\textbf{Learning from Demonstrations}: Learning from Demonstration (LfD) involves teaching agents tasks by observing human or agent demonstrations~\cite{schaal1996learning, argall2009survey}. Approaches include Imitation Learning (IL), where agents directly imitate demonstrated behaviors~\cite{ross2011reduction}, and Inverse Reinforcement Learning (IRL), where agents learn the underlying objectives from demonstrations~\cite{ng2000algorithms}. While LfD has been widely applied in robotics and autonomous systems~\cite{breazeal2002robots, ho2016generative}, its application to web agents is less explored. Web agents share similarities with robots in terms of perception, reasoning, and execution~\cite{veloso2005perception}. This overlap suggests that techniques from LfD could enhance the adaptability of web agents to new websites and domains. Our work explores applying LfD to web agents to improve their performance on unseen environments.

\section{Few-Shot Adaptation with Human Demonstrations}
\label{sec:methods_description}

\begin{figure}
    \centering
    \includegraphics[width=1.0\linewidth]{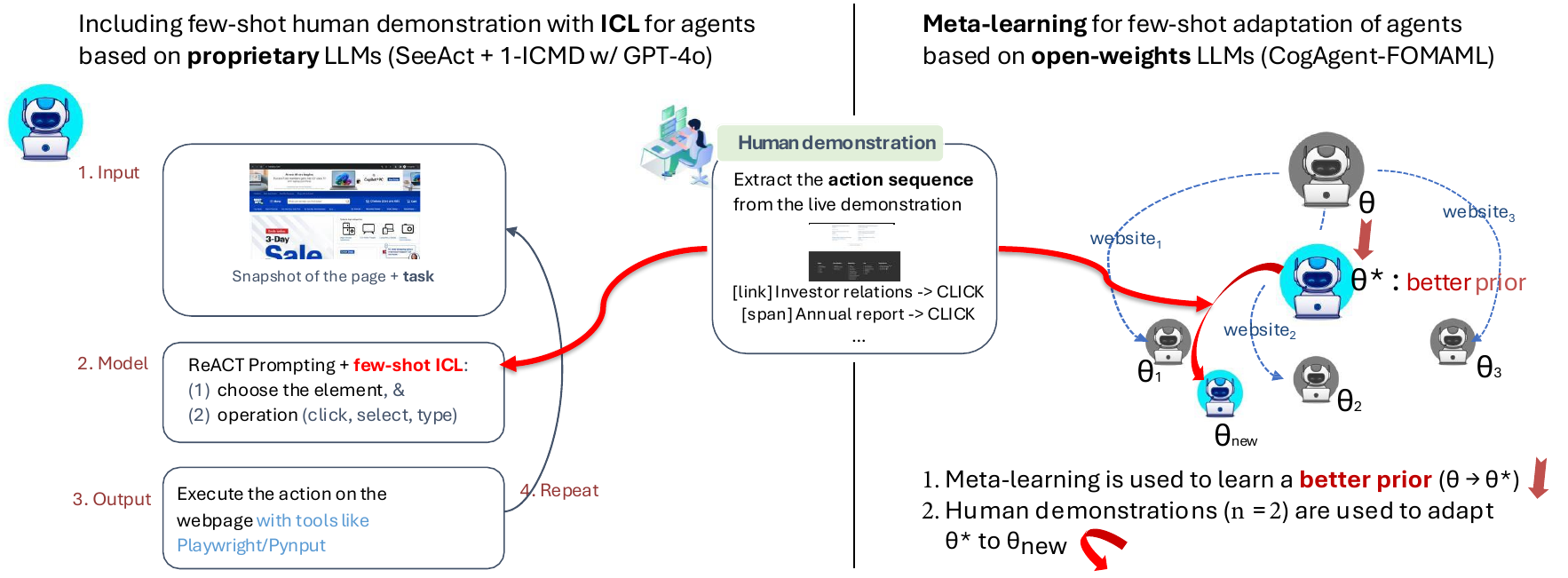}
    \caption{{\textbf{AdaptAgent} for few-shot adaptation of web agents that are based on proprietary and open-weights multimodal LLMs. \textbf{Left}: For proprietary MLLM-based web agents, we include the multimodal human demonstration as in-context examples. \textbf{Right}: For web agents based on open-weights MLLMs, we first learn a better prior using meta-learning and then use few-shot human demonstrations for faster adaptation.
    }}
    \label{fig:methods-overview}
\end{figure}

\noindent\textit{\textbf{Methodological motivation.}} Learning from human demonstrations~\cite{schaal1996learning} has played a key role in many applications, notably helping robots generalize to new tasks or existing tasks under new {environments and } constraints~\cite{argall2009survey}. {Prior work has highlighted the limited generalizability of web agents to unseen tasks, websites, and domains}~\cite{zheng2024seeact, hong2023cogagent}. Agents that automate web tasks and robots that automate real-world tasks share strong analogies in desired capabilities (i.e., perception, reasoning, execution~\cite{veloso2005perception}), allowing for transfer of modeling strategies between these domains. This inspires us to adopt learning from human demonstrations for web agents to {improve their adaptability} to unseen settings. While it’s possible to fine-tune web agents with a large number of human demonstrations covering new websites and domains, such approaches require tedious annotations and are expensive. Therefore, building highly adaptable web agents requires the ability to adapt them in a \textit{data-efficient} manner.

Despite the success of learning from demonstration in adapting robots and the strong analogies between physical robots and web agents, unique challenges remain for web agents. Traditionally, robot learning from human demonstrations {exhibits limited generalizability}; i.e., when a human demonstrates task $\mathcal{A}$ a few times, the robot learns to do the same task $\mathcal{A}$ or closely related tasks, akin to imitation learning~\cite{hussein2017imitation, ren2021generalization}. It remains to be seen how well web agents can generalize to unseen settings with few-shot human demonstrations, which is the primary focus of this work. In other words, can a handful of human demonstrations of {specific} tasks on certain websites (e.g., \textit{"book a flight...''} on \textit{delta.com}) lead the web agent to learn related tasks on {similar} websites (e.g., \textit{"check visa requirements...’’} on \textit{united.com}), or even generalize to unrelated domains (e.g., \textit{``book a driving test appointment...’’} on \textit{dol.wa.gov})? Our work proposes \texttt{AdaptAgent}{, a framework} to enable web agents to adapt with few-shot human demonstrations and evaluates their generalizability to unseen settings.

\noindent\textit{\textbf{Methods for learning with human demonstrations.}} Our proposed framework for adapting  multimodal web agents with few-shot human demonstrations builds on advances in proprietary and open-weights multimodal LLMs. We use SeeAct~\cite{zheng2024seeact}, which employs a carefully crafted prompting strategy with GPT-4o, as a representative proprietary model {baseline} and {adapt it using} multimodal in-context examples. As the representative {baseline for} {SoTA} open-weights models, we use CogAgent~\cite{hong2023cogagent} --- an 18B multimodal LLM with a dedicated visual backbone to process GUI images. Given the success of meta-learning in efficient adaptation, we propose fine-tuning models like CogAgent with meta-learning instead of regular fine-tuning. {See Figure \ref{fig:methods-overview} for an overview of our proposed \texttt{AdaptAgent} framework.} Next, we elaborate on the methodological details for in-context learning and meta-learning with human demonstrations for proprietary and open-weights models, respectively.

\noindent\textit{1. In-context learning with SeeAct + GPT-4o}: SeeAct uses a carefully constructed prompt (using ReAct prompting~\cite{yao2022react}) to {guide multimodal LLMs like GPT-4o in iteratively determining} the next action based on the current GUI state to complete the user-assigned task. In-context learning (ICL) {has proven} to be an effective approach for adapting proprietary LLMs~\cite{bertsch2024context}. Consequently, we deconstruct the human demonstration of a task on the target website/domain into a sequence of 
(visual snapshot, HTML elements (filtered following ~\cite{zheng2024seeact}), human selection) and include them as an ICL example with the original SeeAct prompt; see Appendix \ref{table:icmd_prompt} and Figure \ref{fig:methods-overview} (left).

\noindent\textit{2. Meta-learning with CogAgent}: To overcome the limited abilities of general-purpose multimodal LLMs to process GUI snapshots --- which involve complex layout understanding, OCR, and functional understanding of HTML elements, Hong et al. (2023)~\cite{hong2023cogagent} pre-trained general-purpose MLLMs like CogVLM~\cite{wang2023cogvlm} on tasks involving GUI processing. Beyond extensive pre-training, fine-tuning on task-specific datasets showed notable performance boosts for CogAgent over several baselines. In this work, we consider the pre-trained CogAgent and further adapt it using model-agnostic meta-learning (MAML)~\cite{finn2017modelfirst} with few-shot {human} demonstrations; refer to Figure \ref{fig:methods-overview} (right) for a visual depiction.

Meta-learning~\cite{schmidhuber1987evolutionary}, often dubbed “learning to learn”, is a training strategy {in which a model learns to adapt efficiently to unseen tasks by leveraging knowledge gained from updates across many related tasks.} Model-agnostic meta-learning~\cite{finn2017modelfirst} is one such approach applicable to any model.  Mathematically, the meta-learned model $\theta^*$ is obtained via meta-updates $
\theta \leftarrow \theta - \beta\cdot\nabla_{\theta}\sum_i^N\mathcal{L}_{\mathcal{T}_i}(\theta_i)$ (outer loop update), where $\beta$ is the meta-learning step size, and the gradient is derived from the sum of losses $\mathcal{L}_{\mathcal{T}_i}(\theta_i)$ across all tasks. Each $\theta_i$ is initialized from $\theta$ and fine-tuned on task $\mathcal{T}_i$, via $
\theta_i \leftarrow \theta - \alpha\cdot\nabla_{\theta}\mathcal{L}_{\mathcal{T}_i}(\theta)$ (inner loop update), with $\alpha$ being the step size. Thus, each meta-update involves meta-gradients (gradients through gradients). However, since our experiments involve LLMs with billions of parameters, computing meta-gradients is computationally challenging. Therefore, we consider the first-order approximation of model-agnostic meta-learning (FOMAML). FOMAML has demonstrated performance on par with MAML~\cite{finn2017modelfirst, nichol2018first}, potentially due to the predominantly locally linear nature of neural networks~\cite{goodfellow2014explaining, razzhigaev2024your}, {making the second-order gradients} negligible. Therefore, our meta-learning updates are represented as (derivation in Appendix \ref{ref:fomaml-approximation}): 
$\theta \leftarrow \theta - \beta \cdot \sum_{i=1}^N \nabla_{\theta} \mathcal{L}_{\mathcal{T}_i}(\theta_i)$. In other words, {when} adapting multimodal {web} agents with meta-learning, the inner loop involves fine-tuning the agent ($\theta \rightarrow \theta_i$) on web tasks $\mathcal{T}_i$ from a given website{, with the training subset used for this inner loop denoted as $\mathcal{D}_i^{train}$.} Then, for the outer loop update, we update the parameters of the MLLM agent $\theta$ by backpropagating the gradients of the loss at $\theta_i$, where the loss is computed on held-out web tasks from the same website/domain — denoted as $\mathcal{D}_i^{test}$. Importantly, the gradients being backpropagated are computed at $\theta_i$ {(rather than $\theta$), ensuring the MLLM agent is not trained on both $\mathcal{D}_i^{train}$ \& $\mathcal{D}_i^{test}$.} Essentially, we train the MLLM agent $\theta$ on $\mathcal{D}_i^{train}$ to obtain $\theta_i$ and then update its \textit{original} parameters $\theta$ using penalties {based on how well $\theta_i$ performs on held-out $\mathcal{D}_i^{test}$. A better $\theta^*$ serves as a better starting point to arrive at better $\theta_i$ through fine-tuning, leading to less penalties on held-out $\mathcal{D}_i^{test}$.} This ensures quick and data-efficient adaptation of the agent to unseen websites. 

\section{Experimental Protocol and Details}
\label{sec:experimental_protocal_and_details}

\noindent{\textbf{Datasets:} To evaluate the quick adaptation capabilities of our agents, we design experiments {that require} adaptation to unseen websites and domains. We consider two widely used benchmarks: Mind2Web~\cite{deng2023mind2web} and VisualWebArena~\cite{koh2024visualwebarena}. \textbf{\textit{Mind2Web}} provides standardized train and test sets across various websites and domains. The train set includes {1,009} tasks from {73} websites and {3} domains, while the test set is categorized into cross-task {(174 tasks from 64 seen websites)}, cross-website {(142 tasks from 10 unseen websites)}, and cross-domain {(694 tasks from 2 unseen domains)} subsets to evaluate different aspects of generalization. Since the cross-task evaluation set overlaps with the train set, we propose minor amendments to ensure proper evaluation of adaptability (details in Appendix~\ref{app_sec:benchmark_details}). \textbf{\textit{VisualWebArena}} simulates a live environment with three different websites (Reddit, Classifieds, and Shopping) to evaluate task success rates of web agents. We use the entire VisualWebArena benchmark {(910 tasks)} to test the adaptability of our web agent to unseen websites. While some tasks have step-level ground truth, others provide only an overall task success signal based on the environment’s state. More details about the datasets are presented in Appendix \ref{app_sec:benchmark_details}. 

\textbf{Experimental Protocol}: 
Our experimental protocol for developing and evaluating the adaptability of web agents {varies based} on whether the underlying multimodal LLM is proprietary or open-weights. For the \textbf{proprietary} model (i.e., GPT-4o), we use the prompting method proposed in SeeAct and add one {ICL} example from the website or domain to which the agent should adapt. This ICL example acts as the one-shot {($n=1$)} human demonstration (denoted as 1-ICMD for 1 in-context multimodal demonstration). We adopted a one-shot setting for ICL given the trade-off between time and incremental accuracy improvements; see Appendix \ref{app_sec:effect-of-more-icl}.
The selection of the ICL example ensures relevance to the cross-task, cross-website, and cross-domain evaluation setups. Specifically, for {Mind2Web's} \textit{cross-task} and \textit{cross-website} evaluation, we randomly sample one task from the same website (for cross-task) or from each unique website (for cross-website) in the test set and evaluate on the remaining examples from that website, maintaining website-level correspondence. For \textit{cross-domain} evaluation,  we randomly sample one task from each unique domain in the cross-domain test set and evaluate on the remaining examples from that domain. For \textit{VisualWebArena} evaluation,  we randomly choose one task as the in-context demonstration from the website being evaluated. For the \textbf{open-weights model} (i.e., CogAgent), 
during meta-learning, we sample 4 tasks per website from the 73 websites in the Mind2Web training set: 2 tasks for adaptation {($\mathcal{D}_i^{train}$)} and 2 tasks {($\mathcal{D}_i^{test}$)} (1 from the same website and 1 from a different website within the same domain) for computing the adaptation loss and updating the agent’s parameters as discussed in Section \ref{sec:methods_description}. After meta-learning, the {meta-trained} model adapts to new websites in the cross-website test set by fine-tuning on 2 tasks from each website and then evaluating on the remaining tasks. For cross-domain evaluation, we adapt on 2 tasks from each new domain and evaluate on the rest within that domain; see Figure \ref{fig:experimental-protocol} in the Appendix. We do not perform website adaptation for the cross-task test set, as all websites are seen during meta-learning. For VisualWebArena, we adapt {the meta-trained model on the Mind2Web training set,} using 2 tasks from each of the 3 websites and evaluate on the remaining tasks. To control for the effect of adaptation tasks, we report average results across 5 independent runs with different task selections. Overall, our {approach} involves {meta-}training the model with 292 tasks from Mind2Web (73 websites × 4 tasks) and adapting with 2 demonstrations to new websites/domains. Implementation details are available in Appendix  \ref{app_sec:implementation_details}. We denote our meta-learned and adapted agent as CogAgent-FOMAML.

We compare the performance of adapted agents with existing {SoTA} agents as \textbf{baselines}. For the proprietary model, zero-shot SeeAct + GPT-4o serves as  a baseline. We also include Set-of-Mark prompting (SoM)~\cite{yang2023set, koh2024visualwebarena} in the image input, giving us a slightly augmented baseline that we denote as SeeAct*. For the open-weights model, we consider the pre-trained CogAgent and another variant—CogAgent-FT—that uses conventional fine-tuning on the entire train set of Mind2Web as baselines. Additionally, we consider CogAgent-FT (DE) as another baseline that maintains data equivalence (DE) with the proposed CogAgent-FOMAML method by using the same training subset for conventional fine-tuning. CogAgent-FOMAML and CogAgent-FT (DE) use 292 examples during meta-learning and fine-tuning, respectively, while CogAgent-FT uses $\sim$3.4× as many examples.

\textbf{Evaluation metrics}: For evaluation on the Mind2Web test sets, since the ground-truth human trajectories are available for each task, we compute granular metrics: the accuracy of predicting the correct HTML element {(Ele. Acc.)} to act on; the $F_1$ score of predicting the correct operation {(Op. $F_1$)} such as click, select, type; the percentage of successful steps {(Step SR)} --- requiring correct prediction of the element, the operation, and the optional type/selection text; and the percentage of successful tasks {(Overall SR)}, where task-level success is achieved only if the entire sequence of steps predicted by the agent aligns with the ground-truth human trajectories. For VisualWebArena, since the ground-truth human trajectories are available only for a subset of the data (233 tasks corresponding to the unique templates) and the rest of the tasks have only a task-level success signal within the live environment, we use the overall success rate as the primary metric while also quantifying the granular metrics specifically for the subset of tasks with human trajectories.

\section{Results}
\textbf{\textit{Few-shot human demonstrations unlock complementary gains in agent's performance}}: 
Table \ref{tab:results-combined} compares the baseline and few-shot adapted versions of proprietary (SeeAct, SeeAct*) and open-weights (CogAgent) models on \textit{(a)} the Mind2Web dataset across cross-task, cross-website, and cross-domain evaluation settings, and \textit{(b)} the VisualWebArena dataset across human trajectories and live environment settings. The proprietary models adapt through multimodal in-context demonstration, while CogAgent adapts via meta-learning.  Recall that for CogAgent, we tested two baseline versions: one fine-tuned on the entire Mind2Web train set and another to ensure date-equivalence with CogAgent-FOMAML.

\begin{table*}[ht]
\begin{subtable}{1\textwidth}
\centering
\resizebox{1.0\textwidth}{!}{
\begin{tabular}{llcccccccccccc}

\toprule
\multirow{3}{*}{\textbf{Type}} & \multirow{3}{*}{\textbf{Model}} & \multicolumn{4}{c}{\textbf{Cross-Task}} & \multicolumn{4}{c}{\textbf{Cross-Website}} & \multicolumn{4}{c}{\textbf{Cross-Domain}} \\
\cmidrule(lr{1em}){3-6}
\cmidrule(lr{1em}){7-10}
\cmidrule(lr{1em}){11-14}
                       & & Ele. Acc. & Op. F1 & Step SR & Overall SR & Ele. Acc. & Op. F1 & Step SR & Overall SR & Ele. Acc. & Op. F1 & Step SR & Overall SR \\
\midrule
\multicolumn{14}{l}{\cellcolor{pale7}\textit{Proprietary Models}}  \\
\midrule
Baseline & SeeAct (GPT-4o) & 62.21 & 66.56 & 56.31 & 14.37 & 55.25 & 58.89 & 49.90 & 15.83 & 57.33 & 60.74 & 53.72 & 19.49 \\
\midrule
\textbf{Adapted}
& SeeAct + 1-ICMD & \textbf{66.29} & \textbf{71.61} & \textbf{60.37} & \textbf{19.69} & \textbf{60.32} & \textbf{64.15} & \textbf{53.91} & \textbf{22.46} & \textbf{60.54} & \textbf{62.97} & \textbf{57.40} & \textbf{23.97} \\
\midrule
\midrule

Baseline & SeeAct* (GPT-4o) & 63.75 & 67.68 & 58.60 & 15.38 & 57.02 & 60.01 & 50.05 & 15.89 & 59.30 & 62.80 & 54.82 & 19.88 \\
\midrule
\textbf{Adapted} 
& \cellcolor{pale4}SeeAct* + 1-ICMD & \cellcolor{pale4} \textbf{67.77} & \cellcolor{pale4} \textbf{72.52} & \cellcolor{pale4} \textbf{61.88} & \cellcolor{pale4} \textbf{22.46} & \cellcolor{pale4} \textbf{61.67} & \cellcolor{pale4} \textbf{64.76} & \cellcolor{pale4} \textbf{53.98} & \cellcolor{pale4} \textbf{23.10} & \cellcolor{pale4} \textbf{62.44} & \cellcolor{pale4} \textbf{65.41} & \cellcolor{pale4} \textbf{58.33} & \cellcolor{pale4} \textbf{24.06} \\
\midrule
\multicolumn{14}{l}{\cellcolor{pale7}\textit{Open-weights Models}}  \\
\midrule

\multirow{2}{*}{Baseline} 
& CogAgent-FT & \textbf{59.46} & \textbf{63.15} & \textbf{54.43} & \textbf{13.36} & 53.17 & 57.03 & 47.14 & 12.42 & 61.36 & 62.79 & 55.71 & 15.20 \\
& CogAgent-FT (DE) & 55.17 & 59.87 & 50.25 & 10.43 & 49.46 & 53.17 & 44.27 & 10.10 & 59.51 & 59.06 & 52.20 & 13.28 \\
\midrule
\textbf{Adapted} 
& CogAgent-FOMAML & 59.34 & 62.82 & 53.32 & 11.89 & \textbf{59.49} & \textbf{62.11} & \textbf{55.38} & \textbf{16.96} & \textbf{62.01} & \textbf{63.13} & \textbf{57.29} & \textbf{19.66} \\
\bottomrule
\end{tabular}%
}
\caption{Mind2Web dataset}
\label{mind2web_main_results}
\end{subtable}

\vspace{0.05in}

\begin{subtable}{1\textwidth}
\centering
\resizebox{0.60\textwidth}{!}{
\begin{tabular}{llccccc}
\toprule
\multirow{3}{*}{\textbf{Type}} & \multirow{3}{*}{\textbf{Model}} & \multicolumn{4}{c}{\textbf{Human Trajectories}} & \multicolumn{1}{c}{\textbf{Live Environment}}\\ 
\cmidrule(lr{1em}){3-6}
\cmidrule(lr{1em}){7-7}
                       & & Ele. Acc. & Op. F1 & Step SR & Overall SR & Overall SR \\
\midrule
\multicolumn{7}{l}{\cellcolor{pale7}\textit{Proprietary Models}}  \\
\midrule
Baseline & SeeAct (GPT-4o) & 56.03 & 57.17 & 52.17 & 18.75 & 17.56 \\
\midrule
\textbf{Adapted}
& SeeAct + 1-ICMD & \textbf{59.15} & \textbf{63.18} & \textbf{55.27} & \textbf{22.42} & \textbf{21.36} \\
\midrule
\midrule

Baseline & SeeAct* (GPT-4o) & 57.52 & 59.16 & 53.16 & 18.78 & 18.04 \\
\midrule
\textbf{Adapted}
& \cellcolor{pale4} SeeAct* + 1-ICMD & \cellcolor{pale4} \textbf{61.46} & \cellcolor{pale4} \textbf{64.12} & \cellcolor{pale4} \textbf{56.72} & \cellcolor{pale4} \textbf{23.86} & \cellcolor{pale4} \textbf{23.15} \\
\midrule
\multicolumn{7}{l}{\cellcolor{pale7}\textit{Open-weights Models}}  \\
\midrule
\multirow{2}{*}{Baseline} 
& CogAgent-FT & 52.31 & 55.64 & 48.70 & 08.78 & 06.43 \\
& CogAgent-FT (DE) & 48.62 & 51.71 & 44.81 & 06.81 & 05.11 \\
\midrule
\textbf{Adapted}
& CogAgent-FOMAML & \textbf{57.20} & \textbf{59.14} & \textbf{51.29} & \textbf{11.01} & \textbf{08.47} \\
\bottomrule
\end{tabular}%
}
\caption{VisualWebArena dataset}
\label{visualwebarena_main_results}
\end{subtable}
\caption{{Effect of few-shot adaptation of web agents; all values are percentages. ICMD denotes the multimodal in-context demonstration. FT refers to fine-tuning, DE denotes fine-tuning with data equivalence with respect to our meta-learned models. \textbf{Adapted} models are our proposed methods. \textbf{Bold} indicates best performance, and \colorbox{pale4}{orange} highlight represents the best overall performance. Model size of GPT-4o: 175B; CogAgent: 18B.
}}
\label{tab:results-combined}
\end{table*}

\textbf{We observe that} few-shot adaptation improved the performance of both proprietary and open-weights models across the two datasets and all settings involving adaptation to unseen websites or domains. Specifically, for Mind2Web's cross-website and cross-domain sets, few-shot adaptation using the \texttt{AdaptAgent} framework resulted in an absolute increase in overall success rate ranging from 4.18\% to 7.21\% over the corresponding unadapted counterparts, which corresponds to a relative increase of 21.03\% to 45.40\% over the current state-of-the-art. The trends are consistent across all the models, demonstrating the effectiveness of using only 1 or 2 human demonstrations via \texttt{AdaptAgent}. Similarly, on VisualWebArena, \texttt{AdaptAgent} led to an absolute increase in overall success rate ranging from 3.36\% to 5.11\%, which corresponds to 28.32\% to 65.75\% relative increase over the SoTA approaches.\footnote{{CogAgent-FOMAML outperformed CogAgent-FT (trained on $\sim$3.4× examples than CogAgent-FOMAML) across all tests except for Mind2Web cross-task, where it outperformed CogAgent-FT (DE) trained with data equivalence. This highlights that with an equal amount of training data, our meta-learned CogAgent-FOMAML outperforms the conventionally fine-tuned model as well as demonstrates greater generalizability to unseen tasks. }} In the following sections, we further investigate the advantage of multimodal in-context demonstrations compared to text-only demonstrations, the role of different data selection strategies during meta-learning, and the role of the number of in-context demonstrations used.

\label{app_sec:additional_results}
\textbf{\textit{Multimodal demonstrations are more effective than text-only demonstrations}}:
In our ablation study, we examined the impact of in-context demonstration modalities—specifically text-only versus multimodal—on our top-performing models, SeeAct and SeeAct*. See Figure \ref{fig:additional_analysis_plots} (left) and 

Table \ref{tab:2datasets-textvsmultimodal}. 
\begin{figure}[!h]
    \centering
    \begin{subfigure}[b]{0.32\textwidth}
        \centering
        \includegraphics[scale = 0.23]{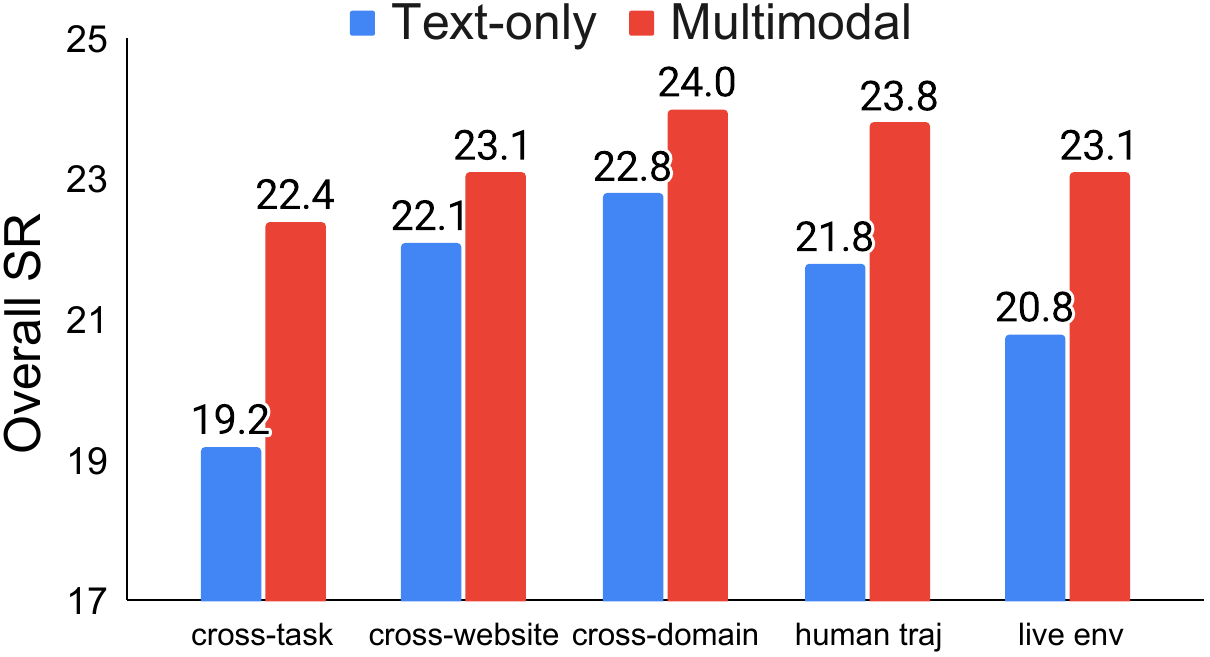}
        \label{fig:plot1}
    \end{subfigure}
    \hfill
    \begin{subfigure}[b]{0.32\textwidth}
        \centering
        \includegraphics[scale = 0.235]{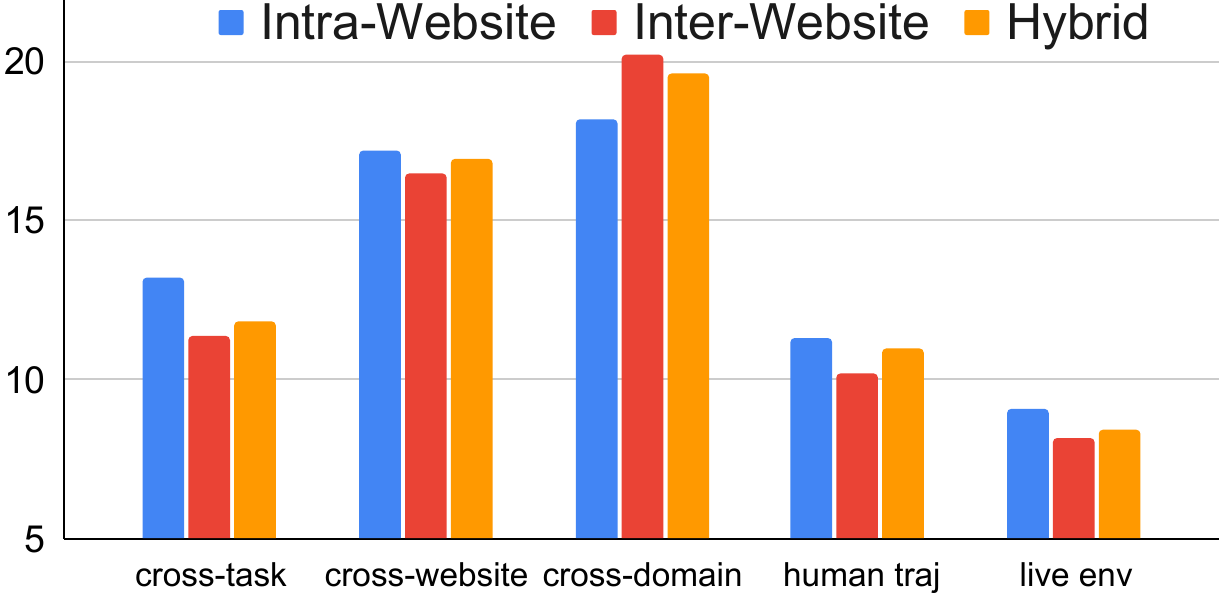}
        \label{fig:plot2}
    \end{subfigure}
    \hfill
    \begin{subfigure}[b]{0.32\textwidth}
        \centering
        \includegraphics[scale = 0.49]{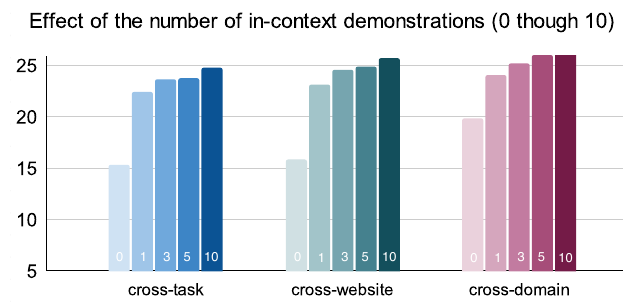}
        \label{fig:plot3}
    \end{subfigure}
    \caption{Additional analyses. \textbf{Left:} Ablation study on demonstration modality in SeeAct*. \textbf{Center:} Comparison of overall SR across meta-learning adaptation strategies in CogAgent. \textbf{Right:} Variation in performance with different numbers of in-context demonstrations; numbers are inset in the bars.}
    \label{fig:additional_analysis_plots}
\end{figure}

\begin{table*}[ht]
\begin{subtable}{1\textwidth}
\centering
\resizebox{1.0\textwidth}{!}{
\begin{tabular}{ll|cccc|cccc|cccc}
\toprule
\multirow{3}{*}{\textbf{Type}} & \multirow{3}{*}{\textbf{Model}} & \multicolumn{4}{c|}{\textbf{Cross-Task}} & \multicolumn{4}{c|}{\textbf{Cross-Website}} & \multicolumn{4}{c}{\textbf{Cross-Domain}} \\
\cmidrule(lr{1em}){3-6}
\cmidrule(lr{1em}){7-10}
\cmidrule(lr{1em}){11-14}
                       & & Ele. Acc. & Op. F1 & Step SR & Overall SR & Ele. Acc. & Op. F1 & Step SR & Overall SR & Ele. Acc. & Op. F1 & Step SR & Overall SR \\
\midrule
Baseline & SeeAct (GPT-4o) & 62.21 & 66.56 & 56.31 & 14.37 & 55.25 & 58.89 & 49.90 & 15.83 & 57.33 & 60.74 & 53.72 & 19.49 \\
\midrule
\multirow{2}{*}{Adapted} & SeeAct + 1-ICTD & 65.71 & 70.82 & 58.19 & 15.91 & 58.94 & 62.87 & 51.11 & 19.56 & 59.31 & 61.69 & 55.23 & 22.16 \\
& SeeAct + 1-ICMD & \textbf{66.29} & \textbf{71.61} & \textbf{60.37} & \textbf{19.69} & \textbf{60.32} & \textbf{64.15} & \textbf{53.91} & \textbf{22.46} & \textbf{60.54} & \textbf{62.97} & \textbf{57.40} & \textbf{23.97} \\
\midrule
\midrule

Baseline & SeeAct* (GPT-4o) & 63.75 & 67.68 & 58.60 & 15.38 & 57.02 & 60.01 & 50.05 & 15.89 & 59.30 & 62.80 & 54.82 & 19.88 \\
\midrule
\multirow{2}{*}{Adapted} & SeeAct* + 1-ICTD & 66.31 & 70.29 & 60.24 & 19.27 & 59.41 & 62.48 & 52.64 & 22.15 & 61.01 & 64.00 & 56.50 & 22.87 \\
& SeeAct* + 1-ICMD & \textbf{67.77} & \textbf{72.52} &  \textbf{61.88} & \textbf{22.46} & \textbf{61.67} & \textbf{64.76} & \textbf{53.98} & \textbf{23.10} & \textbf{62.44} & \textbf{65.41} & \textbf{58.33} &  \textbf{24.06} \\
\bottomrule
\end{tabular}%
}
\caption{Mind2Web dataset}
\label{tab:mind2web-textvsmultimodal}
\end{subtable}

\bigskip
\begin{subtable}{1\textwidth}
\centering
\resizebox{0.7\textwidth}{!}{
\begin{tabular}{ll|cccc|c}
\toprule
\multirow{3}{*}{\textbf{Type}} & \multirow{3}{*}{\textbf{Model}} & \multicolumn{4}{c|}{\textbf{Human Trajectories}} & \multicolumn{1}{c}{\textbf{Live Environment}}\\ 
\cmidrule(lr{1em}){3-6}
\cmidrule(lr{1em}){7-7}
                       & & Ele. Acc. & Op. F1 & Step SR & Overall SR & Overall SR \\
\midrule
Baseline & SeeAct (GPT-4o) & 56.03 & 57.17 & 52.17 & 18.75 & 17.56 \\
\midrule
\multirow{2}{*}{Adapted} & SeeAct + 1-ICTD & 57.16 & 60.74 & 53.92 & 20.56 & 19.12 \\
& SeeAct + 1-ICMD & \textbf{59.15} & \textbf{63.18} & \textbf{55.27} & \textbf{22.42} & \textbf{21.36} \\
\midrule
\midrule

Baseline & SeeAct* (GPT-4o) & 57.52 & 59.16 & 53.16 & 18.78 & 18.04 \\
\midrule
\multirow{2}{*}{Adapted} & SeeAct* + 1-ICTD & 58.98 & 62.93 & 54.54 & 21.82 & 20.87 \\
& SeeAct* + 1-ICMD & \textbf{61.46} & \textbf{64.12} & \textbf{56.72} & \textbf{23.86} & \textbf{23.15} \\
\bottomrule
\end{tabular}%
}
\caption{VisualWebArena dataset}
\label{tab:visualwebarena-textvsmultimodal}
\end{subtable}
\caption{Ablation study on multimodal vs. text-only demonstrations. IC[-]D denotes the type of in-context demonstration, where T and M refer to textual and multimodal demonstrations, respectively. \textbf{Bold} text indicates the best performance for each model.}
\label{tab:2datasets-textvsmultimodal}
\end{table*}
We observe performance improvements with multimodal in-context demonstrations compared to text-only versions. Specifically, across all the settings in the two benchmarks, there was an absolute gain ranging from 0.95\% to 3.78\%, corresponding to a relative increase of 4.29\% and 23.76\%. These results demonstrate the advantage of incorporating richer multimodal in-context demonstrations, including visual snapshots of webpages, compared to relying solely on text.

\textbf{\textit{Data selection strategies for meta-learning influence adaptability in different settings}}: Recall that our implementation of meta-learning uses 2 tasks for the $\theta$ to $\theta_i$ adaptation and 2 other tasks for meta-updates to $\theta$, which eventually leads to $\theta^*$. The selection strategies for these tasks could influence the kind of generalization that the meta-learning encourages after adaptation. For instance, consider the setting where the selection of the 2 tasks for inner-loop adaptation and the 2 tasks for meta-updates is done from the same website. In this \textit{\underline{intra-website}} setup, meta-updates to $\theta$ encourage generalization to within-website tasks after adaptation to few tasks from the same website. However, in an alternate setup, the selection could involve 2 tasks for inner-loop adaptation from website $w_i$ but 2 tasks taken from a different website $w_j : i \neq j$ within the same domain. In this \textit{\underline{inter-website}} strategy, meta-updates to $\theta$ would encourage generalization to other websites within the domain after adaptation to website $w_i$. A \textit{\underline{hybrid}} approach on the other hand, would involve inner-loop adaptation with 2 tasks from website $w_i$ and meta-updates using 2 tasks, of which one is taken from website $w_i$ and the other is taken from a different website $w_j : i \neq j$ within the same domain. Our ablations and empirical results, discussed below, assess how the three data selection strategies influence the agent's performance under different evaluation settings. 

We trained a different variant of CogAgent-FOMAML using each of these three different data selection strategies. Table \ref{tab:2datasets-metalearning-strategies} and Figure \ref{fig:additional_analysis_plots} (center) contrasts the performance of the resulting variants. It is clear that that while the intra-website selections strategy benefits cross-website generalization of the adapted agent, the inter-website strategy is more effective for cross-domain generalization. This trend is consistent across the two benchmarks and aligns with our intuition above. Furthermore, we observe that the hybrid strategy strikes the right balance between generalization across cross-website and cross-domain settings. Therefore, for our main results, we considered the hybrid data selection strategy. Nonetheless, depending on the desired scope of the adapted agent, future research and practitioners could employ a data selection strategy that would be more effective in their setting.  

\begin{table*}[ht]
\begin{subtable}{1\textwidth}
\centering
\resizebox{1.0\textwidth}{!}{
\begin{tabular}{ll|cccc|cccc|cccc}
\toprule
\multirow{3}{*}{\textbf{Type}} & \multirow{3}{*}{\textbf{Model}} & \multicolumn{4}{c|}{\textbf{Cross-Task}} & \multicolumn{4}{c|}{\textbf{Cross-Website}} & \multicolumn{4}{c}{\textbf{Cross-Domain}} \\
\cmidrule(lr{1em}){3-6}
\cmidrule(lr{1em}){7-10}
\cmidrule(lr{1em}){11-14}
                       & & Ele. Acc. & Op. F1 & Step SR & Overall SR & Ele. Acc. & Op. F1 & Step SR & Overall SR & Ele. Acc. & Op. F1 & Step SR & Overall SR \\
\midrule
\multirow{3}{*}{Baseline} & CogAgent & 30.63 & 47.67 & 25.11 & 02.80 & 31.50 & 51.52 & 21.29 & 02.11 & 32.17 & 49.94 & 23.32 & 02.59 \\
& CogAgent-FT & 59.46 & \textbf{63.15} & \textbf{54.43} & \textbf{13.36} & 53.17 & 57.03 & 47.14 & 12.42 & 61.36 & 62.79 & 55.71 & 15.20 \\
& CogAgent-FT (DE with FOMAML) & 55.17 & 59.87 & 50.25 & 10.43 & 49.46 & 53.17 & 44.27 & 10.10 & 59.51 & 59.06 & 52.20 & 13.28 \\
\midrule
\multirow{3}{*}{Adapted} &  CogAgent-FOMAML (intra-website) & \textbf{60.74} & 62.44 & 53.14 & 13.24 & \textbf{60.16} & \textbf{63.47} & \textbf{55.88} & \textbf{17.28} & 61.36 & 62.79 & 55.71 & 18.20 \\
& CogAgent-FOMAML (inter-website) & 58.77 & 62.16 & 53.01 & 11.46 & 59.02 & 62.84 & 54.13 & 16.50 & \textbf{63.88} & \textbf{65.01} & \textbf{58.42} & \textbf{20.22} \\
& CogAgent-FOMAML (hybrid) & 59.34 & 62.82 & 53.32 & 11.89 & 59.49 & 62.11 & 55.38 & 16.96 & 62.01 & 63.13 & 57.29 & 19.66 \\
\bottomrule
\end{tabular}%
}
\caption{Mind2Web dataset}
\label{tab:mind2web-metalearning-strategies}
\end{subtable}

\bigskip
\begin{subtable}{1\textwidth}
\centering
\resizebox{0.8\textwidth}{!}{
\begin{tabular}{ll|cccc|c}
\toprule
\multirow{3}{*}{\textbf{Type}} & \multirow{3}{*}{\textbf{Model}} & \multicolumn{4}{c|}{\textbf{Human Trajectories}} & \multicolumn{1}{c}{\textbf{Live Environment}}\\ 
\cmidrule(lr{1em}){3-6}
\cmidrule(lr{1em}){7-7}
                       & & Ele. Acc. & Op. F1 & Step SR & Overall SR & Overall SR \\
\midrule
\multirow{3}{*}{Baseline}  & CogAgent & 25.27 & 38.64 & 19.61 & 01.31 & 0.46 \\
& CogAgent-FT & 52.31 & 55.64 & 48.70 & 08.78 & 6.43 \\
& CogAgent-FT (DE with FOMAML) & 48.62 & 51.71 & 44.81 & 06.81 & 5.11 \\
\midrule
\multirow{3}{*}{Adapted}  & CogAgent-FOMAML (intra-website) & \textbf{57.36} & \textbf{60.07} & 52.61 & \textbf{11.36} & \textbf{9.17} \\
& CogAgent-FOMAML (inter-website) & 56.11 & 58.44 & \textbf{53.81} & 10.24 & 8.29 \\
& CogAgent-FOMAML (hybrid) & 57.20 & 59.14 & 51.29 & 11.01 & 8.47 \\
\bottomrule
\end{tabular}%
}
\caption{VisualWebArena dataset}
\label{tab:visualwebarena-metalearning-strategies}
\end{subtable}
\caption{Analysis of the three meta-learning adaptation strategies used with the CogAgent model. FT refers to fine-tuning, while DE denotes fine-tuning with data equivalence to the meta-learned models, i.e., using less than one-third of the training data. \textbf{Bold} text indicates the best performance in each evaluation setting.}
\label{tab:2datasets-metalearning-strategies}
\end{table*}

\textbf{\textit{More multimodal demonstrations help boost agent's performance}}: 
\label{app_sec:effect-of-more-icl}
Next, we analyze the impact of increasing the number of in-context multimodal demonstrations on the performance of SeeAct*. Figure \ref{fig:additional_analysis_plots} (right) shows the impact of 1, 3, 5, and 10 in-context multimodal demonstrations on a subset of 30 tasks sampled from the cross-task, cross-website, and cross-domain sets in Mind2Web. Across all the settings, we notice that the performance does improve slightly with more demonstrations. However, the gains are minimal as we add more and more demonstrations. Given the higher computational costs associated with longer prompts and incremental accuracy improvements, it is preferable to utilize a limited number of in-context multimodal demonstrations.

\section{Discussion and Limitations}

We propose the \texttt{AdaptAgent} framework, which uses few-shot human demonstrations for efficient adaptation of web agents to unseen websites and domains, and demonstrated its efficacy for both proprietary and open-weights MLLM-based agents. More broadly, our results indicate that \texttt{AdaptAgent} provides a notable boost in the success rate of current SoTA web agents on unseen websites and domains in a cost-effective way, complementing the gains obtained by building larger pre-trained models or fine-tuning on larger datasets. Beyond the main result, we also demonstrate the benefits of using multimodal in-context demonstrations over text-only demonstrations. Furthermore, our ablations provide actionable recommendations for future research and practitioners to build efficiently adaptable web agents --- \textit{(i)} the trade-offs associated with different data selection strategies for meta-learning can influence the generalizability of the adapted web agent, and (ii) while more in-context multimodal demonstrations boost the performance of proprietary agents, the gains tend to saturate with a higher number of examples.

Despite the state-of-the-art performance achieved by \texttt{AdaptAgent}, the best-performing agent attained an overall task success rate of less than $25\%$ on both Mind2Web and VisualWebArena. There remains significant room for improvement, particularly for tasks requiring long action sequences and websites with complex visual layouts (see \ref{app_sec:difficulty-level-results} for more details), underscoring the potential for future advancements in this area.

\section*{Acknowledgements}
The authors would like to thank Annapoorani Lakshmi Narayanan and Sumitra Ganesh, both with J.P. Morgan AI Research, for valuable discussions and feedback about this work. 

\section*{Disclaimer}
This paper was prepared for informational purposes by the Artificial Intelligence Research group of JPMorgan Chase \& Co and its affiliates ("J.P. Morgan") and is not a product of the Research Department of J.P. Morgan.  J.P. Morgan makes no representation and warranty whatsoever and disclaims all liability, for the completeness, accuracy or reliability of the information contained herein.  This document is not intended as investment research or investment advice, or a recommendation, offer or solicitation for the purchase or sale of any security, financial instrument, financial product or service, or to be used in any way for evaluating the merits of participating in any transaction, and shall not constitute a solicitation under any jurisdiction or to any person, if such solicitation under such jurisdiction or to such person would be unlawful. 

\bibliographystyle{plain}
\bibliography{paper}

\begin{thebibliography}{10}

\bibitem{abbeel2004apprenticeship}
Pieter Abbeel and Andrew~Y Ng.
\newblock Apprenticeship learning via inverse reinforcement learning.
\newblock In {\em Proceedings of the twenty-first international conference on Machine learning}, page~1, 2004.

\bibitem{achiam2023gpt}
Josh Achiam, Steven Adler, Sandhini Agarwal, Lama Ahmad, Ilge Akkaya, Florencia~Leoni Aleman, Diogo Almeida, Janko Altenschmidt, Sam Altman, Shyamal Anadkat, et~al.
\newblock Gpt-4 technical report.
\newblock {\em arXiv preprint arXiv:2303.08774}, 2023.

\bibitem{argall2009survey}
Brenna~D Argall, Sonia Chernova, Manuela Veloso, and Brett Browning.
\newblock A survey of robot learning from demonstration.
\newblock {\em Robotics and autonomous systems}, 57(5):469--483, 2009.

\bibitem{bertsch2024context}
Amanda Bertsch, Maor Ivgi, Uri Alon, Jonathan Berant, Matthew~R Gormley, and Graham Neubig.
\newblock In-context learning with long-context models: An in-depth exploration.
\newblock {\em arXiv preprint arXiv:2405.00200}, 2024.

\bibitem{boisvert2024workarena}
L{\'e}o Boisvert, Megh Thakkar, Maxime Gasse, Massimo Caccia, De~Chezelles, Thibault Le~Sellier, Quentin Cappart, Nicolas Chapados, Alexandre Lacoste, and Alexandre Drouin.
\newblock Workarena++: Towards compositional planning and reasoning-based common knowledge work tasks.
\newblock {\em arXiv preprint arXiv:2407.05291}, 2024.

\bibitem{breazeal2002robots}
Cynthia Breazeal and Brian Scassellati.
\newblock Robots that imitate humans.
\newblock {\em Trends in cognitive sciences}, 6(11):481--487, 2002.

\bibitem{brown2019extrapolating}
Daniel Brown, Wonjoon Goo, Prabhat Nagarajan, and Scott Niekum.
\newblock Extrapolating beyond suboptimal demonstrations via inverse reinforcement learning from observations.
\newblock In {\em International conference on machine learning}, pages 783--792. PMLR, 2019.

\bibitem{brown2020language}
Tom~B Brown.
\newblock Language models are few-shot learners.
\newblock {\em arXiv preprint arXiv:2005.14165}, 2020.

\bibitem{calinon2010learning}
Sylvain Calinon, Florent D'halluin, Eric~L Sauser, Darwin~G Caldwell, and Aude~G Billard.
\newblock Learning and reproduction of gestures by imitation.
\newblock {\em IEEE Robotics \& Automation Magazine}, 17(2):44--54, 2010.

\bibitem{calinon2007learning}
Sylvain Calinon, Florent Guenter, and Aude Billard.
\newblock On learning, representing, and generalizing a task in a humanoid robot.
\newblock {\em IEEE Transactions on Systems, Man, and Cybernetics, Part B (Cybernetics)}, 37(2):286--298, 2007.

\bibitem{chen2021learning}
Letian Chen, Rohan Paleja, and Matthew Gombolay.
\newblock Learning from suboptimal demonstration via self-supervised reward regression.
\newblock In {\em Conference on robot learning}, pages 1262--1277. PMLR, 2021.

\bibitem{cheng2024seeclick}
Kanzhi Cheng, Qiushi Sun, Yougang Chu, Fangzhi Xu, Yantao Li, Jianbing Zhang, and Zhiyong Wu.
\newblock Seeclick: Harnessing gui grounding for advanced visual gui agents.
\newblock {\em arXiv preprint arXiv:2401.10935}, 2024.

\bibitem{das2021model}
Neha Das, Sarah Bechtle, Todor Davchev, Dinesh Jayaraman, Akshara Rai, and Franziska Meier.
\newblock Model-based inverse reinforcement learning from visual demonstrations.
\newblock In {\em Conference on Robot Learning}, pages 1930--1942. PMLR, 2021.

\bibitem{deng2023mind2web}
Xiang Deng, Yu~Gu, Boyuan Zheng, Shijie Chen, Samuel Stevens, Boshi Wang, Huan Sun, and Yu~Su.
\newblock Mind2web: Towards a generalist agent for the web, 2023.

\bibitem{englert2017inverse}
Peter Englert, Ngo~Anh Vien, and Marc Toussaint.
\newblock Inverse kkt: Learning cost functions of manipulation tasks from demonstrations.
\newblock {\em The International Journal of Robotics Research}, 36(13-14):1474--1488, 2017.

\bibitem{finn2017modelfirst}
Chelsea Finn, Pieter Abbeel, and Sergey Levine.
\newblock Model-agnostic meta-learning for fast adaptation of deep networks.
\newblock In {\em International conference on machine learning}, pages 1126--1135. PMLR, 2017.

\bibitem{finn2017one}
Chelsea Finn, Tianhe Yu, Tianhao Zhang, Pieter Abbeel, and Sergey Levine.
\newblock One-shot visual imitation learning via meta-learning.
\newblock In {\em Conference on robot learning}, pages 357--368. PMLR, 2017.

\bibitem{goodfellow2014explaining}
Ian~J Goodfellow, Jonathon Shlens, and Christian Szegedy.
\newblock Explaining and harnessing adversarial examples.
\newblock {\em arXiv preprint arXiv:1412.6572}, 2014.

\bibitem{gur24webagent}
Izzeddin Gur, Hiroki Furuta, Austin Huang, Mustafa Safdari, Yutaka Matsuo, Douglas Eck, and Aleksandra Faust.
\newblock A real-world webagent with planning, long context understanding, and program synthesis, 2024.

\bibitem{he2024webvoyager}
Hongliang He, Wenlin Yao, Kaixin Ma, Wenhao Yu, Yong Dai, Hongming Zhang, Zhenzhong Lan, and Dong Yu.
\newblock Webvoyager: Building an end-to-end web agent with large multimodal models.
\newblock {\em arXiv preprint arXiv:2401.13919}, 2024.

\bibitem{ho2016generative}
Jonathan Ho and Stefano Ermon.
\newblock Generative adversarial imitation learning.
\newblock {\em Advances in neural information processing systems}, 29, 2016.

\bibitem{hong2023cogagent}
Wenyi Hong, Weihan Wang, Qingsong Lv, Jiazheng Xu, Wenmeng Yu, Junhui Ji, Yan Wang, Zihan Wang, Yuxiao Dong, Ming Ding, and Jie Tang.
\newblock Cogagent: A visual language model for gui agents, 2023.

\bibitem{humphreys2022data}
Peter~C Humphreys, David Raposo, Tobias Pohlen, Gregory Thornton, Rachita Chhaparia, Alistair Muldal, Josh Abramson, Petko Georgiev, Adam Santoro, and Timothy Lillicrap.
\newblock A data-driven approach for learning to control computers.
\newblock In {\em International Conference on Machine Learning}, pages 9466--9482. PMLR, 2022.

\bibitem{hussein2017imitation}
Ahmed Hussein, Mohamed~Medhat Gaber, Eyad Elyan, and Chrisina Jayne.
\newblock Imitation learning: A survey of learning methods.
\newblock {\em ACM Computing Surveys (CSUR)}, 50(2):1--35, 2017.

\bibitem{iki2022berts}
Taichi Iki and Akiko Aizawa.
\newblock Do berts learn to use browser user interface? exploring multi-step tasks with unified vision-and-language berts.
\newblock {\em arXiv preprint arXiv:2203.07828}, 2022.

\bibitem{jin2023dataefficientalignmentlargelanguage}
Di~Jin, Shikib Mehri, Devamanyu Hazarika, Aishwarya Padmakumar, Sungjin Lee, Yang Liu, and Mahdi Namazifar.
\newblock Data-efficient alignment of large language models with human feedback through natural language, 2023.

\bibitem{khattab2023dspy}
Omar Khattab, Arnav Singhvi, Paridhi Maheshwari, Zhiyuan Zhang, Keshav Santhanam, Sri Vardhamanan, Saiful Haq, Ashutosh Sharma, Thomas~T Joshi, Hanna Moazam, et~al.
\newblock Dspy: Compiling declarative language model calls into self-improving pipelines.
\newblock {\em arXiv preprint arXiv:2310.03714}, 2023.

\bibitem{koh2024visualwebarena}
Jing~Yu Koh, Robert Lo, Lawrence Jang, Vikram Duvvur, Ming~Chong Lim, Po-Yu Huang, Graham Neubig, Shuyan Zhou, Ruslan Salakhutdinov, and Daniel Fried.
\newblock Visualwebarena: Evaluating multimodal agents on realistic visual web tasks.
\newblock {\em arXiv preprint arXiv:2401.13649}, 2024.

\bibitem{kuderer2015learning}
Markus Kuderer, Shilpa Gulati, and Wolfram Burgard.
\newblock Learning driving styles for autonomous vehicles from demonstration.
\newblock In {\em 2015 IEEE international conference on robotics and automation (ICRA)}, pages 2641--2646. IEEE, 2015.

\bibitem{lai2024autowebglm}
Hanyu Lai, Xiao Liu, Iat~Long Iong, Shuntian Yao, Yuxuan Chen, Pengbo Shen, Hao Yu, Hanchen Zhang, Xiaohan Zhang, Yuxiao Dong, et~al.
\newblock Autowebglm: Bootstrap and reinforce a large language model-based web navigating agent.
\newblock {\em arXiv preprint arXiv:2404.03648}, 2024.

\bibitem{liu2018reinforcement}
Evan~Zheran Liu, Kelvin Guu, Panupong Pasupat, Tianlin Shi, and Percy Liang.
\newblock Reinforcement learning on web interfaces using workflow-guided exploration.
\newblock {\em arXiv preprint arXiv:1802.08802}, 2018.

\bibitem{liu2024deita}
Wei Liu, Weihao Zeng, Keqing He, Yong Jiang, and Junxian He.
\newblock What makes good data for alignment? a comprehensive study of automatic data selection in instruction tuning, 2024.

\bibitem{ng2000algorithms}
Andrew~Y Ng, Stuart Russell, et~al.
\newblock Algorithms for inverse reinforcement learning.
\newblock In {\em Icml}, volume~1, page~2, 2000.

\bibitem{nichol2018first}
A~Nichol.
\newblock On first-order meta-learning algorithms.
\newblock {\em arXiv preprint arXiv:1803.02999}, 2018.

\bibitem{noy2023experimental}
Shakked Noy and Whitney Zhang.
\newblock Experimental evidence on the productivity effects of generative artificial intelligence.
\newblock {\em Science}, 381(6654):187--192, 2023.

\bibitem{oracle_ai_agents_2024}
{Oracle}.
\newblock Oracle {AI} agents help organizations achieve new levels of productivity, Sep 2024.
\newblock Online press release.

\bibitem{ou2024synatra}
Tianyue Ou, Frank~F Xu, Aman Madaan, Jiarui Liu, Robert Lo, Abishek Sridhar, Sudipta Sengupta, Dan Roth, Graham Neubig, and Shuyan Zhou.
\newblock Synatra: Turning indirect knowledge into direct demonstrations for digital agents at scale.
\newblock {\em arXiv preprint arXiv:2409.15637}, 2024.

\bibitem{ouyang2022training}
Long Ouyang, Jeffrey Wu, Xu~Jiang, Diogo Almeida, Carroll Wainwright, Pamela Mishkin, Chong Zhang, Sandhini Agarwal, Katarina Slama, Alex Ray, et~al.
\newblock Training language models to follow instructions with human feedback.
\newblock {\em Advances in neural information processing systems}, 35:27730--27744, 2022.

\bibitem{pomerleau1988alvinn}
Dean~A Pomerleau.
\newblock Alvinn: An autonomous land vehicle in a neural network.
\newblock {\em Advances in neural information processing systems}, 1, 1988.

\bibitem{rafailov2024direct}
Rafael Rafailov, Archit Sharma, Eric Mitchell, Christopher~D Manning, Stefano Ermon, and Chelsea Finn.
\newblock Direct preference optimization: Your language model is secretly a reward model.
\newblock {\em Advances in Neural Information Processing Systems}, 36, 2024.

\bibitem{ravichandar2020recent}
Harish Ravichandar, Athanasios~S Polydoros, Sonia Chernova, and Aude Billard.
\newblock Recent advances in robot learning from demonstration.
\newblock {\em Annual review of control, robotics, and autonomous systems}, 3:297--330, 2020.

\bibitem{razzhigaev2024your}
Anton Razzhigaev, Matvey Mikhalchuk, Elizaveta Goncharova, Nikolai Gerasimenko, Ivan Oseledets, Denis Dimitrov, and Andrey Kuznetsov.
\newblock Your transformer is secretly linear.
\newblock {\em arXiv preprint arXiv:2405.12250}, 2024.

\bibitem{ren2021generalization}
Allen Ren, Sushant Veer, and Anirudha Majumdar.
\newblock Generalization guarantees for imitation learning.
\newblock In {\em Conference on Robot Learning}, pages 1426--1442. PMLR, 2021.

\bibitem{ross2011reduction}
St{\'e}phane Ross, Geoffrey Gordon, and Drew Bagnell.
\newblock A reduction of imitation learning and structured prediction to no-regret online learning.
\newblock In {\em Proceedings of the fourteenth international conference on artificial intelligence and statistics}, pages 627--635. JMLR Workshop and Conference Proceedings, 2011.

\bibitem{rybski2007interactive}
Paul~E Rybski, Kevin Yoon, Jeremy Stolarz, and Manuela~M Veloso.
\newblock Interactive robot task training through dialog and demonstration.
\newblock In {\em Proceedings of the ACM/IEEE international conference on Human-robot interaction}, pages 49--56, 2007.

\bibitem{schaal1996learning}
Stefan Schaal.
\newblock Learning from demonstration.
\newblock {\em Advances in neural information processing systems}, 9, 1996.

\bibitem{schaal2006dynamic}
Stefan Schaal.
\newblock Dynamic movement primitives-a framework for motor control in humans and humanoid robotics.
\newblock In {\em Adaptive motion of animals and machines}, pages 261--280. Springer, 2006.

\bibitem{schmidhuber1987evolutionary}
Jurgen Schmidhuber.
\newblock Evolutionary principles in self-referential learning.
\newblock {\em On learning how to learn: The meta-meta-... hook.) Diploma thesis, Institut f. Informatik, Tech. Univ. Munich}, 1(2):48, 1987.

\bibitem{sclar2023quantifying}
Melanie Sclar, Yejin Choi, Yulia Tsvetkov, and Alane Suhr.
\newblock Quantifying language models' sensitivity to spurious features in prompt design or: How i learned to start worrying about prompt formatting.
\newblock {\em arXiv preprint arXiv:2310.11324}, 2023.

\bibitem{shaikh2024show}
Omar Shaikh, Michelle Lam, Joey Hejna, Yijia Shao, Michael Bernstein, and Diyi Yang.
\newblock Show, don't tell: Aligning language models with demonstrated feedback.
\newblock {\em arXiv preprint arXiv:2406.00888}, 2024.

\bibitem{shaw2023pixels}
Peter Shaw, Mandar Joshi, James Cohan, Jonathan Berant, Panupong Pasupat, Hexiang Hu, Urvashi Khandelwal, Kenton Lee, and Kristina Toutanova.
\newblock From pixels to ui actions: Learning to follow instructions via graphical user interfaces, 2023.

\bibitem{shi2017world}
Tianlin Shi, Andrej Karpathy, Linxi Fan, Jonathan Hernandez, and Percy Liang.
\newblock World of bits: An open-domain platform for web-based agents.
\newblock In {\em International Conference on Machine Learning}, pages 3135--3144. PMLR, 2017.

\bibitem{silver2016mastering}
David Silver, Aja Huang, Chris~J Maddison, Arthur Guez, Laurent Sifre, George Van Den~Driessche, Julian Schrittwieser, Ioannis Antonoglou, Veda Panneershelvam, Marc Lanctot, et~al.
\newblock Mastering the game of go with deep neural networks and tree search.
\newblock {\em nature}, 529(7587):484--489, 2016.

\bibitem{song2024beyond}
Yueqi Song, Frank Xu, Shuyan Zhou, and Graham Neubig.
\newblock Beyond browsing: Api-based web agents.
\newblock {\em arXiv preprint arXiv:2410.16464}, 2024.

\bibitem{veloso2005perception}
Manuela Veloso.
\newblock Perception, cognition, and action in teams of robots.
\newblock Colloquium at the Department of Computer Science, Princeton University, September 28 2005.

\bibitem{wang2023cogvlm}
Weihan Wang, Qingsong Lv, Wenmeng Yu, Wenyi Hong, Ji~Qi, Yan Wang, Junhui Ji, Zhuoyi Yang, Lei Zhao, Xixuan Song, et~al.
\newblock Cogvlm: Visual expert for pretrained language models.
\newblock {\em arXiv preprint arXiv:2311.03079}, 2023.

\bibitem{wolf2020transformers}
Thomas Wolf, Lysandre Debut, Victor Sanh, Julien Chaumond, Clement Delangue, Anthony Moi, Pierric Cistac, Tim Rault, R{\'e}mi Louf, Morgan Funtowicz, et~al.
\newblock Transformers: State-of-the-art natural language processing.
\newblock In {\em Proceedings of the 2020 conference on empirical methods in natural language processing: system demonstrations}, pages 38--45, 2020.

\bibitem{yang2023set}
Jianwei Yang, Hao Zhang, Feng Li, Xueyan Zou, Chunyuan Li, and Jianfeng Gao.
\newblock Set-of-mark prompting unleashes extraordinary visual grounding in gpt-4v.
\newblock {\em arXiv preprint arXiv:2310.11441}, 2023.

\bibitem{yao2022react}
Shunyu Yao, Jeffrey Zhao, Dian Yu, Nan Du, Izhak Shafran, Karthik Narasimhan, and Yuan Cao.
\newblock React: Synergizing reasoning and acting in language models.
\newblock {\em arXiv preprint arXiv:2210.03629}, 2022.

\bibitem{zhang2022selfd}
Jimuyang Zhang, Ruizhao Zhu, and Eshed Ohn-Bar.
\newblock Selfd: self-learning large-scale driving policies from the web.
\newblock In {\em Proceedings of the IEEE/CVF Conference on Computer Vision and Pattern Recognition}, pages 17316--17326, 2022.

\bibitem{zhao2023group}
Siyan Zhao, John Dang, and Aditya Grover.
\newblock Group preference optimization: Few-shot alignment of large language models.
\newblock {\em arXiv preprint arXiv:2310.11523}, 2023.

\bibitem{zheng2024seeact}
Boyuan Zheng, Boyu Gou, Jihyung Kil, Huan Sun, and Yu~Su.
\newblock Gpt-4v(ision) is a generalist web agent, if grounded.
\newblock In {\em Forty-first International Conference on Machine Learning}, 2024.

\bibitem{ziebart2008maximum}
Brian~D Ziebart, Andrew~L Maas, J~Andrew Bagnell, Anind~K Dey, et~al.
\newblock Maximum entropy inverse reinforcement learning.
\newblock In {\em Aaai}, volume~8, pages 1433--1438. Chicago, IL, USA, 2008.

\end{thebibliography}

\appendix
\section{Appendix}

\subsection{Detailed Related Work}
\label{sec:detailed_related_work}
\subsubsection{UI/Web Agents}

AI-enabled digital device control~\cite{shi2017world, humphreys2022data} --- i.e., controlling digital devices using AI with natural language as input --- has been a long-standing ambition for large-scale automation of inherently useful tasks. The underlying problem involves mapping a language instruction from the user to a sequence of digital actions that AI agents can execute to successfully complete the task. Before LLMs, approaches to the problem involved using \textit{reinforcement learning} on top of (often pre-trained) language models like LSTM and BERT for processing language input and HTML/DOM along with ResNet-like models for processing GUI states~\cite{humphreys2022data, liu2018reinforcement, iki2022berts}.  More recently, as multimodal LLMs have demonstrated success in modeling vision-and-language, they have lent themselves as strong backbones for building web agents that can process tasks specified by the user and engage in reasoning to output the best possible actions to be executed on a user interface such as a web browser. A majority of SoTA work \cite{zheng2024seeact,he2024webvoyager} use a pretrained, off-the-shelf LLM such as GPT-4(V/o) to build such multimodal web agents. The input information being provided as context to the LLM can include an image of the GUI, a series of prior actions, additional overlaid image annotations, as well as the HTML/DOM information assuming that the task is web interaction and access to HTML/DOM is possible.  Work such as Pix2Act~\cite{shaw2023pixels} and WebAgent \cite{gur24webagent} train LLMs to attend to parts of HTML code or generate the next action step through self-supervision, or combine the effectiveness of MLLMs with the promise of reinforcement learning train agents via  Behavioral cloning or REINFORCE. However, these works were usually trained on simpler sandboxed environments and would require significant training resources as well as tedious curation of data samples \cite{lai2024autowebglm}.  A disadvantage of such an approach is that it cannot scale to tasks that are complex or  that use proprietary enterprise software. Additionally, agents that require exploration as part of the training process would need constant human supervision to avoid risky outcomes. While there has been considerable progress in the success rate of agents on tasks that are encountered as part of their training, their performance in unseen settings has been lacking. To the best of our knowledge, prior work has not explicitly focused on methods that could make Web/GUI agents more adaptable to unseen settings. 

Our work proposes a framework where GUI/web agents are trained to efficiently adapt to unseen settings using few-shot human demonstrations. Data-efficient adaptation of web agents via human demonstrations will (a) avoid costly retraining processes/updates for unseen settings, (b) boost the generalizability of web agents to complex workflows and proprietary settings, and (c) enable web agents to learn from custom information provided by human experts as a part of the demonstrations. 

\subsubsection{Few-shot learning with LLMs} 

Data-efficient alignment of LLMs to preferences and new tasks is an active area of research~\cite{jin2023dataefficientalignmentlargelanguage,liu2024deita}. In contrast to relatively data-hungry approaches like RLHF~\cite{ouyang2022training} and DPO~\cite{rafailov2024direct} that often require hundreds of thousands paired comparisons, few-shot alignment and adaptation aims to use a limited number of examples. While in-context learning~\cite{brown2020language} is one of the approaches to enable few-shot adaptation of LLMs, it is known to be tedious~\cite{khattab2023dspy} and is sensitive to variations~\cite{sclar2023quantifying}. Fine-tuning alternatives, like GPO~\cite{zhao2023group} and DITTO~\cite{shaikh2024show} have shown promises in few-shot tuning an LLM to align to subjective preferences demonstrated in tasks like email writing and opinion-based question-answering. Most notably, ~\cite{zhao2023group} proposes Group Preference Optimization (GPO), which is a meta-learning framework to update LLM parameters based on few-shot in-context demonstrations. However, it is unclear if few-shot alignment approaches like GPO and DITTO, designed for subjective preference tuning, translate to more concrete predictive tasks like ours. Nonetheless, the broader motivation behind methods like GPO -- i.e., meta-learning, is a promising opportunity to improve the performance of multimodal web agents, especially cross-website and cross-domains scenarios. Inspired by the promise of meta learning and learning from demonstrations, we adopt model-agnostic meta-learning~\cite{finn2017modelfirst} to train web agents to adapt quickly.

\subsubsection{Learning from demonstrations}
Learning from Demonstration (LfD)~\cite{schaal1996learning, breazeal2002robots, argall2009survey, ravichandar2020recent} involves teaching agents tasks by observing human or agent demonstrations, enabling them to acquire skills by either directly imitating actions in supervised learning settings~\cite{ross2011reduction} or using demonstrations as guidance in reinforcement learning settings~\cite{abbeel2004apprenticeship}. This approach helps agents master complex tasks that are difficult to explicitly program.

The two main approaches to LfD are Imitation Learning (IL) and Inverse Reinforcement Learning (IRL). Imitation Learning (IL) centers on the direct imitation of demonstrated expert behaviors, where agents replicate observed actions using methods like Behavioral Cloning~\cite{pomerleau1988alvinn}, and DAgger (Dataset Aggregation)~\cite{ross2011reduction}. IL typically involves mapping human demonstrations to agent actions through supervised learning. Early techniques such as Dynamic Movement Primitives (DMPs)~\cite{schaal2006dynamic} encoded movement trajectories, while probabilistic models like Gaussian Mixture Models (GMMs)~\cite{calinon2007learning} and Hidden Markov Models (HMMs)~\cite{calinon2010learning} captured variability and intent in demonstrations. However, IL has limitations when learning from suboptimal demonstrations, as it focuses on mimicking behavior rather than understanding the underlying objectives. Inverse Reinforcement Learning (IRL), in contrast, seeks to uncover the underlying objective of the task by learning a reward function from demonstrations~\cite{englert2017inverse, brown2019extrapolating, chen2021learning, das2021model}. Instead of merely imitating behavior, IRL infers the goal the demonstrator is optimizing. Once the reward function is learned, Reinforcement Learning (RL) can be used to autonomously derive a policy that achieves the task's goal, allowing the agent to explore and optimize its actions beyond the initial demonstrations~\cite{ng2000algorithms}. Some notable extensions of IRL include apprenticeship learning~\cite{abbeel2004apprenticeship}, maximum entropy IRL~\cite{ziebart2008maximum}, and generative adversarial imitation learning (GAIL)~\cite{ho2016generative}. Applications of LfD span robotics, enabling adaptation to various environments and objects~\cite{breazeal2002robots, rybski2007interactive, argall2009survey}; autonomous driving, where vehicles learn navigation and decision-making from human driving data~\cite{kuderer2015learning, zhang2022selfd}; and game playing, including chess and Go, where agents replicate human gameplay~\cite{silver2016mastering}. 

Agents that automate web tasks share significant similarities with robots that perform real-world tasks, as both rely on core capabilities like perception, reasoning, and execution~\cite{veloso2005perception}. This overlap enables the transfer of modeling techniques between the two areas. Drawing on this analogy, our work explores applying learning from human demonstrations to web agents to enhance their adaptability on unseen websites and domains.

\subsection{First-order approximation of MAML for Multimodal Web Agents}
\label{ref:fomaml-approximation}
We present a derivation of the first-order approximation of MAML proposed by~\cite{finn2017modelfirst}, while contextualizing it to our setting of updating multimodal LLMs.  We begin with the original expression for updates using the MAML algorithm in Equation \ref{eq:maml-expression}:

\begin{equation}
        \theta \leftarrow \theta - \beta \cdot \nabla_{\theta} \sum_{i=1}^N \mathcal{L}_{\mathcal{T}_i}(\theta_i).
\label{eq:maml-expression}
\end{equation}
Using the chain rule, the derivative term can be expressed as 
$\sum_{i=1}^N (\nabla_{\theta}\theta_i \times \nabla_{\theta_i}\mathcal{L}_{\mathcal{T}_i}(\theta_i))$.
The first component within the summation could be broken down further as,
\[
    \nabla_{\theta} \theta_i = \nabla_{\theta} (\theta - \alpha \cdot \nabla_{\theta}\mathcal{L}^{train}(\theta)),
\]
where $\mathcal{L}^{train}$ denotes the loss on the examples used for training $\theta_i$ from task $\mathcal{T}_i$ and $\alpha$ denotes the step-size in the inner loop of meta-training. The above equation further simplifies to
\[
    \nabla_{\theta} \theta_i = \mathbf{I} - \alpha \cdot \nabla_{\theta}^2\mathcal{L}^{train}(\theta).
\]
Now, assuming the second-order derivatives in the expression to zero, provides $\nabla_{\theta}\theta_i = \mathbf{I}$. Plugging that in the original MAML expression gives,

\[
    \theta \leftarrow \theta - \beta \cdot \sum_{i=1}^N \nabla_{\theta_i} \mathcal{L}_{\mathcal{T}_i}(\theta_i).
\]

In our context, this essentially means that the inner loop of meta-learning involves fine-tuning the MLLM agent (i.e., $\theta \rightarrow \theta_i$) on web tasks $\mathcal{T}_i$ from a given website. Let's denote this subset of tasks used for the inner loop of training as $\mathcal{D}_i^{train}$. Following this, we update the parameters of the MLLM agent $\theta$ by back-propagating the gradients of the loss at $\theta_i$, where the loss is computed on held-out web tasks from the same website --- denoted as $\mathcal{D}_i^{test}$. It is worth emphasizing that the gradients being back-propagated are computed at $\theta_i$ (as opposed to $\theta$, which would have resulted in training the MLLM agent on $\mathcal{D}_i^{train}$ and $\mathcal{D}_i^{test}$). In other words, we train the MLLM agent $\theta$ on $\mathcal{D}_i^{train}$ to obtain $\theta_i$ and then update its \textit{original} parameters $\theta$ using penalties computed by evaluating how far $\theta_i$ is from the ``ideal answers'' on held-out $\mathcal{D}_i^{test}$. If exposed to enough updates over varying-but-related websites $i \in \{1, \ldots, N\}$, the updates to the MLLM agent $\theta$ would position it such that it would learn to adapt to unseen websites \textit{quickly} in a data-efficient manner.

\subsection{Benchmark Details}
\label{app_sec:benchmark_details}
\subsubsection{Mind2Web}
\textbf{Training Set}: The training set of the Mind2Web benchmark comprises 1,009 task instances spanning 73 websites from three domains: \textit{travel}, \textit{entertainment}, and \textit{shopping}. These tasks involve various user goals such as booking flights, purchasing tickets, and shopping for products. Each task is accompanied by detailed annotations, including the user instruction, the sequence of actions required to complete the task, and the corresponding HTML and visual states of the web pages.

\textbf{Test Set}: The test set is divided into three subsets to facilitate the evaluation of models in different generalization scenarios:\\
\textit{Cross-Task Subset}: This subset contains 174 tasks from the 64 websites that are present in the training set. The tasks are different from those in the training set but occur on familiar websites and within the same domains.\\
\textit{Cross-Website Subset}: This subset includes 142 tasks from 10 websites that are entirely unseen during training. The websites belong to the same domains as those in the training set.\\
\textit{Cross-Domain Subset}: This subset consists of 694 tasks spanning 53 websites from two new domains: \textit{information} and \textit{service}. These domains are not present in the training set, and the websites are entirely new to the agent.

\textbf{Fixing overlaps between the train and cross-task evaluation sets of Mind2Web}: It is important to note that the standardized cross-task evaluation set of Mind2Web exhibits substantial overlap with the tasks in the training set, which could potentially inflate the evaluation results by testing on tasks that are not truly unseen. For instance, when we computed Jaccard similarity (i.e., intersection-over-union of unique unigrams) between all the tasks in the standardized Mind2Web train set and the cross-task test set, we found pairs of highly similar tasks spread across the two sets. E.g., ``add Prometheus movie to watchlist.'' (train set) and ``add The Wire to the watchlist.'' (cross-task set); ``find a cheapest flight from London to New York on 9th May.'' (train set) and ``find cheapest flight from New York to Toronto, Canada on 29 April.'' (cross-task set). To address this issue, we first combined all the tasks within the existing train and cross-task subsets of the Mind2Web benchmark and computed pair-wise Jaccard similarity between all tasks belonging to the same website. For each website, we then moved $K$ tasks that exhibited least maximum similarity with any other task from the website to construct the amended cross-task evaluation set, while keeping the rest of the tasks from the website in the amended train set. The value of $K$ was determined so as to ensure that the amended train and cross-task sets had the same number of data points as the original train and cross-task sets. We also qualitatively inspected the overlap between the amended train cross-task sets and found that even the most similar tasks (based on unigram Jaccard similarity) across the two sets were now considerable different. For e.g., ``show me all the events at any six flags park in Texas'' (amended train) and ``show me all the artists with smith in their name'' (amended cross-task); ``add to my cart a women's T-shirt priced under 10 dollars'' (amended train) and ``list Batman collectible figures priced under 10 dollars and a customer rating above 4 with a same-day delivery option''  (amended cross-task). This simple-but-important amendment to the Mind2Web's train and cross-task set ensures minimal overlap between tasks seen during training of the web agents and tasks that they are evaluated on in the cross-task setting. 

\subsubsection{VisualWebArena}
The VisualWebArena benchmark comprises 910 tasks representing 233 unique task templates spread across the three websites. Out of the 910 tasks, 233 tasks (one for each task template) have step-level ground truth available in the form of human trajectories. These trajectories provide detailed action sequences that a human would take to accomplish the task, serving as a reference for evaluating the agent’s performance at each step. The remaining tasks do not have step-level ground truth but provide an overall task success signal based on the live environment’s state after the agent’s interaction.

\begin{figure}
    \centering
    \includegraphics[width=1.0\linewidth]{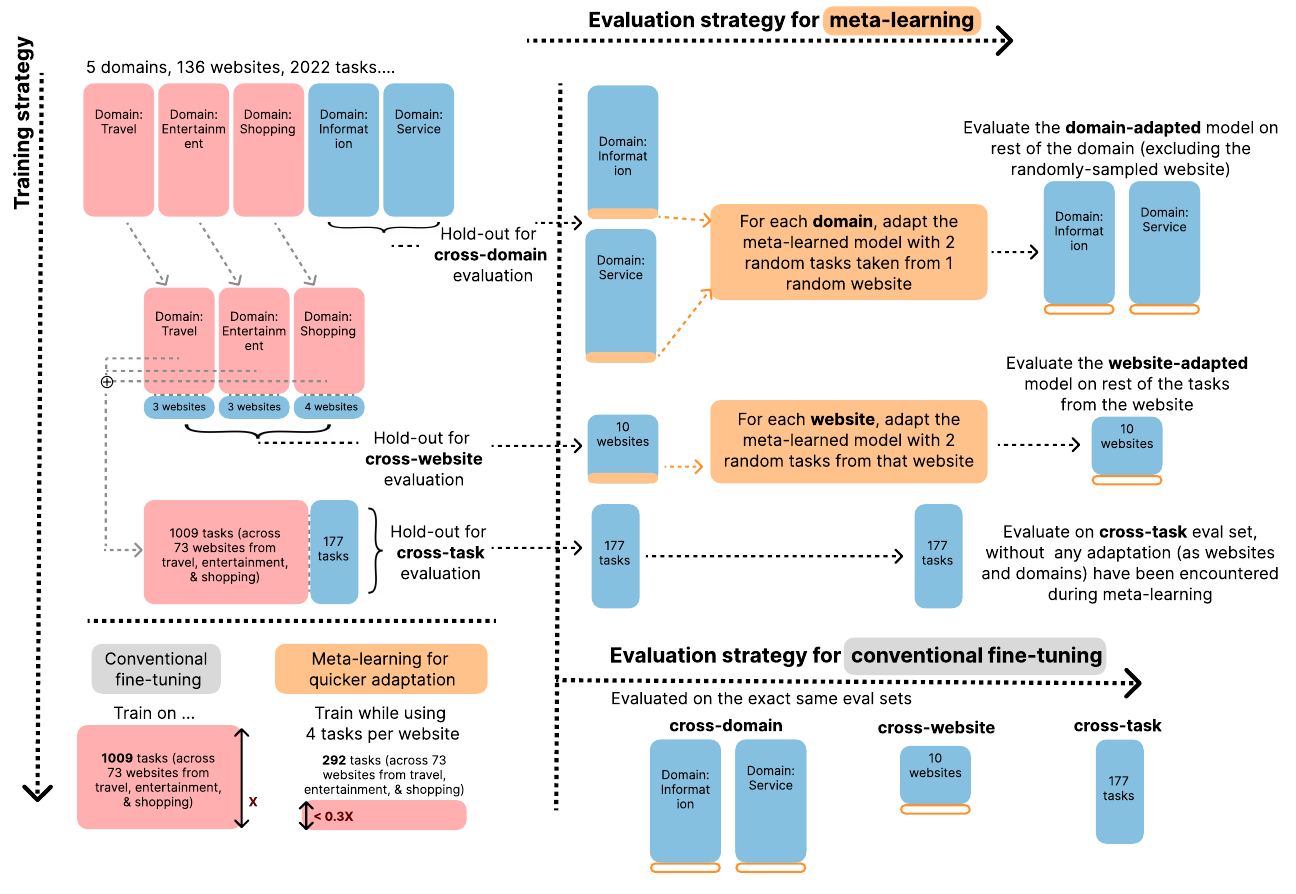}
    \caption{Visual depiction of the protocol used for meta-learning using the Mind2Web train set (left), and the meta-adaptation done on cross-domain and cross-website evaluation sets (top-right). For completeness, we also show the conventional fine-tuning strategy (bottom-right).}
    \label{fig:experimental-protocol}
\end{figure}

\subsection{Implementation Details}
\label{app_sec:implementation_details}
The specific prompts used for experimenting with SeeAct variants, including the modifications to include (text-only/multimodal) in-context demonstrations are presented in Appendix \ref{table:icmd_prompt}. We filtered the top-50 HTML elements to be included in the prompt using the methods adopted by Deng et al.~\cite{deng2023mind2web} and Zeng et al. \cite{zheng2024seeact}.  For experiments with CogAgent, we use the \texttt{THUDM/cogagent-chat-hf} model on HuggingFace~\cite{wolf2020transformers} as the pre-trained version. For updating the model parameters during fine-tuning, meta-learning, and adaptation, we adopted Low-Rank Adaptation (LoRA) with following hyper-parameters: rank $\alpha$ of 20 and learning rate of 1e-5. For fine-tuning, we trained the model for 2 epochs, with other hyper-parameters set to default/the values used by Hong et al. ~\cite{hong2023cogagent}. For meta-learning, we used a meta-batch size of 1, meaning that we trained the agent to adapt to 1 website during the inner-loop, and used one gradient optimization step for each step of the 2 tasks used for loss computation within the inner-loop. For adaptation to new websites and domains, we use the same strategy to adopt one gradient step optimization per step of the 2 sampled tasks to maintain consistency with the training regime. All the experiments were performed on a  virtual server with 8 NVIDIA L4 GPUs (24GiB each). 

\subsection{Results stratified by sequence and visual difficulty levels}
\label{app_sec:difficulty-level-results}
Next, we study the variation of overall SR across difficulty levels, stratified based on (1) sequence complexity; and (2) visual difficulty. The three levels of difficulty in both cases and datasets are easy, medium, and hard, following the protocol established in VisualWebArena.

\noindent\textbullet \hspace{1pt} Sequence difficulty is determined by the length of the ground-truth action sequence (i.e., $\leq3$: easy; $4-9$: medium; $\geq10$: hard). 

\noindent\textbullet \hspace{1pt} 
To assign visual difficulty labels in Mind2Web based on the required visual processing, we used in-context learning with GPT-4o, utilizing labeled VisualWebArena samples as in-context examples. Snapshots of webpages were evaluated as action sequences and categorized as easy, medium, or hard.
Three rounds of annotation were conducted to estimate the self-consistency of GPT annotations, employing chain-of-thought (CoT) reasoning in each round. Finally, human validation was performed to assess the consistency and reasoning of the annotations, with less than 5\% of the total examples having their labels changed based on human review.

Table \ref{tab:results-combined-stratified} compares the baseline and adapted overall SR of SeeAct* and CogAgent, stratified by difficulty (easy, medium, hard) across sequence complexity and visual difficulty in Mind2Web and VisualWebArena settings. 
\begin{table*}[ht]
\centering
\resizebox{1.0\textwidth}{!}{
\begin{tabular}{llc|c|c||c|c}
\toprule
\multirow{4}{*}{\textbf{Type}} & \multirow{4}{*}{\textbf{Model}} & \multicolumn{3}{c}{\textbf{Mind2Web}}  & \multicolumn{2}{c}{\textbf{VisualWebArena}} \\
\cmidrule(lr{1em}){3-5}
\cmidrule(lr{1em}){6-7}
& & \textbf{Cross-Task} & \textbf{Cross-Website} & \textbf{Cross-Domain} & \textbf{Human Trajectories} & \textbf{\textbf{Live Environment}} \\ 
& & Overall SR    & Overall SR   & Overall SR & Overall SR   & Overall SR   \\
& & Easy | Medium | Hard    & Easy | Medium | Hard    & Easy | Medium | Hard  & Easy | Medium | Hard    & Easy | Medium | Hard    \\
\midrule
\multirow{3}{*}{Baseline} & SeeAct* (GPT-4o)    & 15.38 & 15.89 & 19.88 & 18.78 & 18.04\\
& $\hookrightarrow$ Sequence complexity & 56.7\% | 13.7\% | 0.0\% &  57.5\% | 14.1\% | 0.6\% & 58.2\% | 16.5\% | 2.9\% & 57.5\% | 15.2\% | 1.7\% & 56.3\% | 14.6\% | 0.9\%\\
& $\hookrightarrow$ Visual difficulty   & 26.8\% | 11.2\% | 0.0\% & 27.1\% | 11.7\% | 0.9\% & 31.6\% | 13.6\% | 2.6\% & 30.5\% | 12.7\% | 1.6\% & 29.4\% | 11.7\% | 0.8\% \\
\midrule

\multirow{3}{*}{Adapted} & SeeAct* + 1-ICMD    & 22.46 & 23.10 & 24.06  & 23.86 & 23.15 \\
& $\hookrightarrow$ Sequence complexity & 61.3\% | 18.8\% | 1.7\% & 62.6\% | 19.2\% | 2.5\% & 63.6\% | 21.7\% | 5.8\% & 62.6\% | 20.4\% | 5.2\% & 61.6\% | 19.3\% | 5.9\%\\
& $\hookrightarrow$ Visual difficulty   & 33.1\% | 16.2\% | 0.3\% & 33.8\% | 16.6\% | 1.4\%& 36.2\% | 18.4\% | 4.2\% & 35.3\% | 16.9\% | 4.8\% & 34.8\% | 14.9\% | 4.1\% \\
\midrule
\midrule

\multirow{3}{*}{Baseline} & CogAgent-FT (DE) & 10.43 & 10.10 & 13.28 & 06.81 & 5.11\\
& $\hookrightarrow$ Sequence complexity & 38.5\% | 9.3\% | 0.0\% & 36.5\% | 9.0\% | 0.4\% & 39.7\% | 11.2\% | 2.0\% & 20.9\% | 5.5\% | 0.6\% & 15.9\% | 4.1\% | 0.3\%\\
& $\hookrightarrow$ Visual difficulty   & 18.2\% | 7.6\% | 0.0\% & 17.2\% | 7.4\% | 0.6\% & 21.5\% | 09.3\% | 1.8\% & 11.1\% | 4.6\% | 0.6\% & 08.3\% | 3.3\% | 0.2\%\\
\midrule
\multirow{3}{*}{Adapted} & CogAgent-FOMAML & 11.89 & 16.96 & 19.66 &  11.01 & 8.47\\
& $\hookrightarrow$ Sequence complexity & 43.9\% | 10.6\% | 0.6\% & 43.8\% | 10.8\% | 0.7\% & 50.4\% | 14.2\% | 2.5\% & 26.0\% | 6.8\% | 0.8\% & 19.3\% | 5.0\% | 0.4\%\\
& $\hookrightarrow$ Visual difficulty   & 20.7\% | 08.7\% | 0.3\% & 20.6\% | 08.9\% | 0.7\% & 27.3\% | 11.8\% | 2.3\% & 13.8\% | 5.7\% | 0.7\% & 11.5\% | 3.9\% | 0.3\%\\
\bottomrule
\end{tabular}%
}
\caption{Adaptation results stratified by sequence complexity and visual difficulty levels.}
\label{tab:results-combined-stratified}
\end{table*}

We observe that the improvements in adaptation persist when stratified by different difficulty levels, with adaptation enhancing performance across all sequence and visual difficulty levels. SeeAct*, with 1-shot multimodal demonstration, performs best across all difficulty levels. Overall SR decreases as difficulty increases across all model variations, aligning with expectations. The adapted SeeAct* performed better overall, particularly on hard tasks (in terms of both visual and sequence difficulty) in the Mind2Web cross-website and cross-domain evaluation settings, as well as in both VisualWebArena evaluation settings. It showed even greater improvement on tasks with high sequence difficulty compared to those with high visual difficulty. For example, in VisualWebArena, for tasks with hard sequence complexity, overall SR increased from 1.7\% to 5.2\% in human trajectory evaluation and from 0.9\% to 5.9\% in live environment evaluation. In contrast, the gains with the adapted version of CogAgent were minimal on hard tasks, especially in the VisualWebArena evaluation settings.

\subsection{1-ICMD Prompt for SeeAct and SeeAct*}
\label{table:icmd_prompt}
In our approach, we extend the prompt design from \cite{zheng2024seeact} by adding an in-context multimodal demonstration (ICMD). The prompt provided to the GPT-4o model is as follows:

\begin{tcolorbox}[colback=gray!5,colframe=gray!80,title=In-Context Multimodal Demonstration, width=1.0\textwidth, height=75mm]
\small{
\textit{(... preceded by the SeeAct prompt...)}\\
To begin with, here is a quick example of one of the many tasks you could be performing on the website \texttt{\textbf{<website\_name>}}.

Example task's description: \texttt{\textbf{<task\_description>}}

To do this task, you could take the steps shown below.

\medskip

<\textbf{Image} depicting the GUI snapshot at this stage>\\
\textbf{ELEMENT:} \texttt{<element\_name\_1>}

\textbf{ACTION:} \texttt{<action\_type\_1>}

\textbf{VALUE:} \texttt{<value\_if\_applicable\_1>}

\medskip

<\textbf{Image} depicting the GUI snapshot at this stage>\\
\textbf{ELEMENT:} \texttt{<element\_name\_2>}

\textbf{ACTION:} \texttt{<action\_type\_2>}

\textbf{VALUE:} \texttt{<value\_if\_applicable\_2>}

\medskip

$\cdots$

\medskip

This marks the end of an example task and its steps. Now, let's move on to the task at hand.\\
\textit{(... followed by the SeeAct prompt...)}
}
\end{tcolorbox}

\end{document}